\documentclass[10pt,twocolumn,twoside]{IEEEtran} 
\ifCLASSINFOpdf
\else
\fi
\def \submission {}
\usepackage{color}
\usepackage{ulem}
\usepackage{cite}
\usepackage{amssymb,amsmath}
\usepackage{multirow}
\usepackage{caption}
\usepackage{ctable}
\usepackage{array}
\usepackage{hyperref}
\usepackage{graphicx,subfigure}

\definecolor{crimson}{rgb}{0.86, 0.08, 0.24}
\definecolor{tabG}{rgb}{0, 0.5, 0.25}
\definecolor{DarkGreen}{rgb}{0.00, 0.40, 0.00}
\definecolor{RoyalBlue}{rgb}{0.15, 0.25, 0.54}
\definecolor{DarkCyan}{rgb}{0.0, 0.54, 0.54}
\ifx \submission \undefined

\newcommand{\djc}[1]{\textcolor{DarkGreen}{#1}}

\else

\newcommand{\djc}[1]{{#1}}

\fi

\begin{document}
%
\title{SwipeCut: Interactive Segmentation \\ with Diversified Seed Proposals}
%
%
%

\author{Ding-Jie~Chen, Hwann-Tzong~Chen, and~Long-Wen~Chang 
\thanks{D.-J.~Chen is a Ph.D. student at the Department of Computer Science, National Tsing Hua University, Hsinchu 300, Taiwan (e-mail:dj$\_$chen$\_$tw@yahoo.com.tw).}
\thanks{H.-T.~Chen and L.-W.~Chang are with the Department of Computer Science, National Tsing Hua University, Hsinchu 300, Taiwan, and also with the Institute of Information Systems and Applications, National Tsing Hua University, Hsinchu 300, Taiwan (e-mail: \{htchen, lchang\}@cs.nthu.edu.tw).}
\thanks{Manuscript received Month Day, Year; revised Month Day, Year.}}

\maketitle

\begin{abstract}
Interactive image segmentation algorithms rely on the user to provide annotations as the guidance.
When the task of interactive segmentation is performed on a \textbf{small touchscreen device}, the requirement of providing precise annotations could be cumbersome to the user.
We design an efficient seed proposal method that actively proposes annotation seeds for the user to label. The user only needs to check which ones of the query seeds are inside the region of interest (ROI).
We enforce the sparsity and diversity criteria on the selection of the query seeds. At each round of interaction the user is only presented with a small number of informative query seeds that are far apart from each other. As a result, we are able to derive a user friendly interaction mechanism for annotation on small touchscreen devices. The user merely has to \emph{swipe} through on the ROI-relevant query seeds, which should be easy since those gestures are commonly used on a touchscreen.
The performance of our algorithm is evaluated on six publicly available datasets. The evaluation results show that our algorithm achieves high segmentation accuracy, with short response time and less user feedback.
\end{abstract}


%
\IEEEpeerreviewmaketitle

\vspace{-5mm}
\section{Introduction}
\begin{figure}[t]
    \centering
    \subfigure[] { \includegraphics[height=0.11\textwidth]{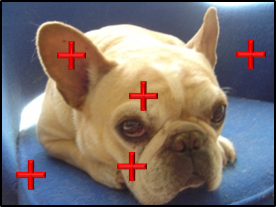} } 
    \subfigure[] { \includegraphics[height=0.11\textwidth]{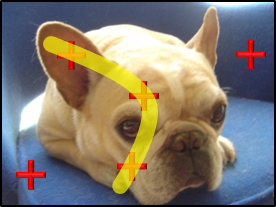} }  
    \subfigure[] { \includegraphics[height=0.11\textwidth]    
{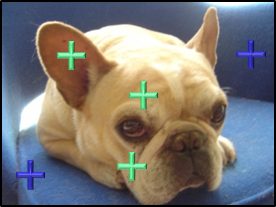} } 
    \subfigure[] { \includegraphics[height=0.11\textwidth]{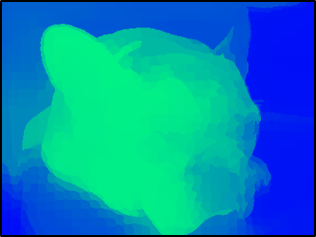} }
    \vspace{-4mm}
    \caption{\label{fig:teaser}
    A label acquiring process of the SwipeCut algorithm. We present a seed proposal algorithm for assisting the user in an interactive image segmentation scenario. The algorithm singles out a few informative seeds as the queries so that the user only needs to \emph{swipe} through the relevant query seeds to specify the region of interest.
    (a) An input image with five scattered crosses (in red) generated by our algorithm as the query seeds.
    (b) The user's gesture (in yellow) swipes through the relevant query seeds that are inside the region of interest.
    (c) The collected seeds of ROI labels (in green) and non-ROI labels (in blue) according to the user's gesture.
    (d) The result of label propagation according to the labels in (c).  }
\end{figure}

%
%
%
%
\IEEEPARstart{I}{mage} segmentation is not a trivial task, especially for images that contain multiple objects and cluttered backgrounds. Interactive image segmentation, or image segmentation with human in the loop, can make the region of interest more clearly defined for obtaining accurate segmentation. Popular interactive image segmentation algorithms allow users to guide the segmentation with some feedback, \djc{in forms of seeds or line-drawings \cite{BoykovJ01,DongSSY15,FengPCC16,Grady06,GulshanRCBZ10,WangHC14,XuPCYH16,LiewWXOF17,ManinisCTG18}, contours \cite{KassWT88,MortensenB95,XianZCXD16,BadoualSUU17,XuPCYH17}, bounding boxes \cite{ChengPZTR15,RotherKB04,XuPCYH17},} or queries \cite{ChenCC16,RupprechtPN15}. In this article, we propose a novel interaction mechanism for acquiring labels via swipe gestures, see Fig.~\ref{fig:teaser} for illustration. The main novelty is that, with the new mechanism, the user does not need to annotate meticulously to prevent crossing the region boundaries while specifying the region of interest. It is particularly suitable for imprecise input interface such as touchscreen.

\begin{figure}[t]
\centering
    \subfigure[] { \includegraphics[height=0.11\textwidth]{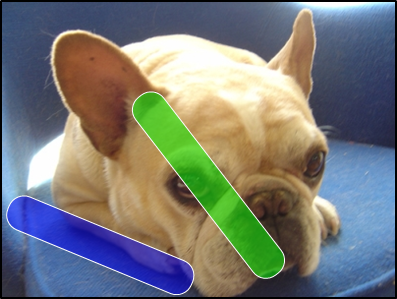} } 
    \subfigure[] { \includegraphics[height=0.11\textwidth]{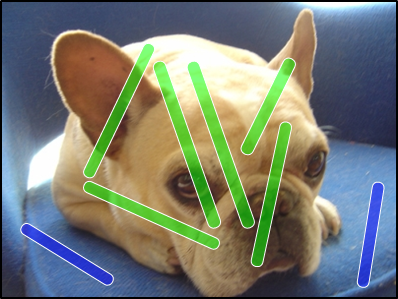} }
    \vspace{-3mm}
    \caption{\label{fig:failLabels}
    Examples of under-qualified annotations. Green strokes are ROI annotations and blue strokes are non-ROI annotations.
    (a) The constrained interface might cause the blue strokes to straddle the object boundary, which would confuse a segmentation algorithm. Examples of the constrained interface are smart phones or tablets since the small touchscreen devices are inconvenient to draw subtle scribbles with high precision using fingers.
    (b) Unfamiliarity of the segmentation algorithm might result in redundant annotations like the green strokes. Increasing the number of ROI strokes on the dog does not improve much the segmentation accuracy. Adding a non-ROI stroke on the top corners of the image in this case would be more helpful for segmentation. }
\end{figure}

Consider the task of segmenting a given image to achieve acceptable segmentation accuracy. Three main factors may affect the overall processing time of the entire interactive segmentation process. The first factor is the algorithm's \emph{segmentation effectiveness} with respect to the user feedback. The second factor is the algorithm's \emph{response time} for completing one round of segmentation in the loop. The third factor is \emph{under-qualified annotations} during interaction. User annotations could be of insufficient quality due to the constraint of the user interface or the unfamiliarity with the segmentation algorithm, see Fig.~\ref{fig:failLabels}. Both the two cases increase the required number of interactions to revise the user annotations.
In addition, a careful user might intend to avoid bad annotations and thus spends more time to finish the segmentation. Therefore, under-qualified annotations could be the time bottleneck of the entire interactive segmentation process. Most existing approaches consider the first two factors to make the segmentation algorithms effective (the first factor) and efficient (the second factor) under the assumption that qualified user-annotations are easy to obtain (the third factor). Our approach addresses the third factor to help the users unambiguously and effortlessly label the query seeds and thus can reduce the turnaround time for interaction. Notice that, some previous works propose error-tolerant segmentation algorithms for handling erroneous scribbles \cite{BaiW14,SubrPSK13}. However, our approach attempts to prevent under-qualified annotations being generated from the very beginning.

Using fingers to manipulate small touchscreen devices means every stroke contains many pixels: ``The average width of the index finger for most adults translates to about 45-57 pixels\footnote{\url{http://mashable.com/2013/01/03/tablet-friendly-website/}}.''
While it is inconvenient to draw scribbles with high precision on a touchscreen, swiping through a specified point on the touchscreen, in contrast, is much simpler. 
Hence, for small touchscreen devices, an interactive mechanism that proposes a few pixels for user to assign binary labels is more accessible. This fact motivates us to design an interaction mechanism tailored for segmenting images on the small touchscreen devices. The mechanism aims to propose sparse pixels for acquiring the ROI labels or non-ROI labels, according to whether the user swipes through the sparse pixels or not. Since we only care about the proposed pixels being touched or not, we allow the finger to pass through other irrelevant pixels. This kind of interaction greatly reduces the chance of annotating wrong labels. A labeling example of the proposed algorithm is shown in Fig.~\ref{fig:teaser}.

\begin{figure*}[t]
    \centering
    \subfigure[] { \includegraphics[height=0.22\textwidth]{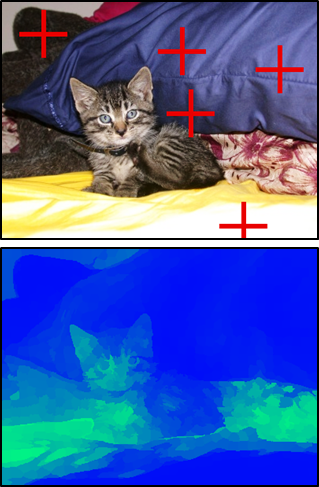} }
    \subfigure[] { \includegraphics[height=0.22\textwidth]{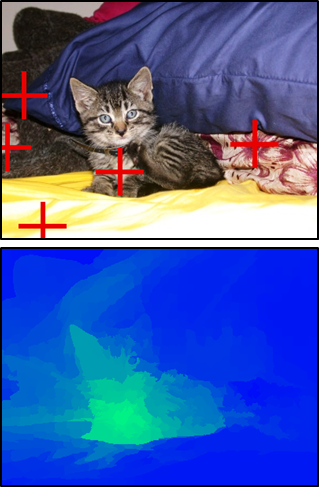} }
    \subfigure[] { \includegraphics[height=0.22\textwidth]{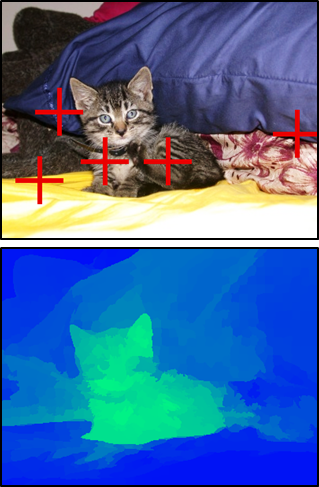} }
    \subfigure[] { \includegraphics[height=0.22\textwidth]{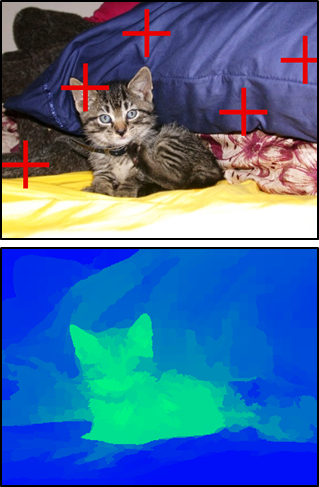} }
    \subfigure[] { \includegraphics[height=0.22\textwidth]{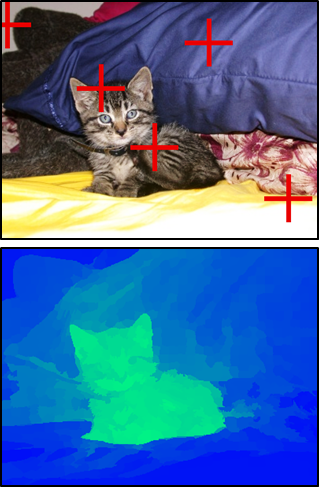} }
    \subfigure[] { \includegraphics[height=0.22\textwidth]{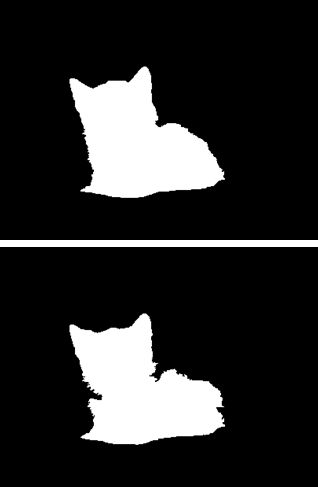} }
    \vspace{-3mm}
    \caption{\label{fig:illustration} A running example of our approach. In (a)-(e), the top images show the scattered crosses (in red) corresponding to five query seeds per round, and the bottom images show the corresponding results of label propagation. (a) The result of the first round. (b) The third round. (c) The fifth round. (d) The seventh round. (e) The tenth round. (f) The ground truth (top) and the segmentation result of the 30th round (bottom).}
\end{figure*}

\begin{figure*}[t]
    \centering
    \includegraphics[width=0.86\textwidth]{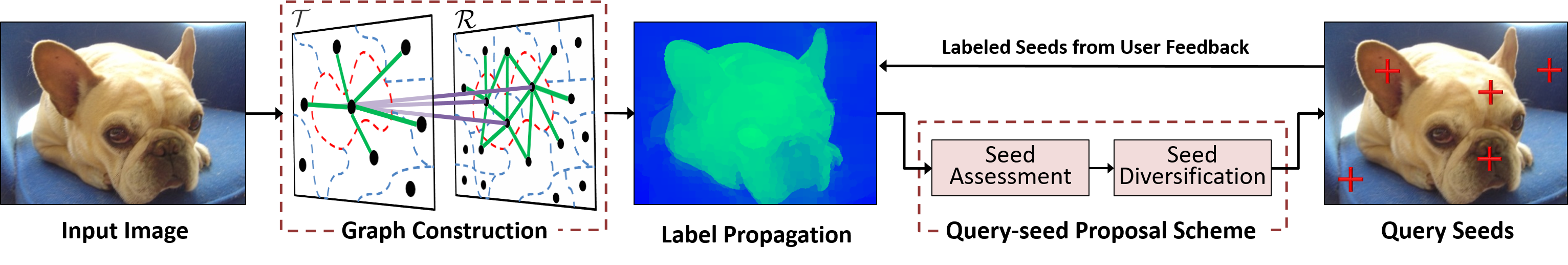}
    \vspace{-3mm}
    \caption{An overview of our interactive image segmentation algorithm. Each input image is first represented by a weighted two-layer graph. At each interaction, we select the first $\theta_k$ top-scored and diversified seeds to acquire true labels from the user. Then, the labeled seeds are used to propagate the labels to all other unlabeled vertices for generating the corresponding segmentation. The next interaction can then be launched with new query seeds derived from the information of label propagation.}
    \label{fig:overview}
\end{figure*}

To implement the novel interaction mechanism for segmenting images on small touchscreen devices, we first propose an effective query-seed proposal scheme built upon a two-layer graph. The two-layer graph consists of moderate-granularity vertices and large-granularity vertices, in which the large-granularity vertices formulate a higher-order soft constraint to make the covered moderate-granularity vertices tend to have the same label. We then diversify the seeds to make them sparse enough in spatial domain for swiping with finger. After acquiring the labels from user's swipe gesture, we propagate the labels to all vertices via calculating the graph distance on the two-layer graph, and hence obtain the segmentation result. 
One running example and an overview of the proposed approach are shown in Fig.~\ref{fig:illustration} and Fig.~\ref{fig:overview}, respectively.

The contributions and advantages of this work are summarized as follows:
\begin{enumerate}[]
\item The interaction mechanism of proposing multiple seeds for user labeling via \emph{swipe gestures}, which is tailored to small touchscreen devices, is new to the image segmentation problem.
\item The user is able to \emph{annotate unambiguously} via swiping through the seeds, since our approach makes the query seeds sparse and separated far enough from each other. 
\item Our method is effective and efficient owing to the proposed informative query seeds selection and label propagation, where both are improved with our higher-order two-layer graph that entails the label consistency.
\item The proposed approach has \emph{high flexibility} for the devices with different touchscreen sizes. The number of query seeds can be adjusted according to the touchscreen size, and hence the additional operations such as zoom-in/zoom-out or drag-and-drop is not needed.
\end{enumerate}

The rest of this article is organized as follows.
Section II reviews related methods on interactive image segmentation and object proposal generation.
Section III formulates the problem to be addressed.
Section IV introduces the proposed interactive image segmentation algorithm, SwipeCut.
Section V shows the experimental results.
Section VI concludes this paper.


\section{Related Work}
We roughly divide interactive image segmentation methods into two categories according to their interaction models: \emph{direct interactive image segmentation} and \emph{indirect interactive image segmentation}. We also refer to several proposal generation methods since some of the ideas and principles are shared with our algorithm.

\paragraph{Direct interactive image segmentation.}
\djc{Many well-known interactive image segmentation algorithms are in this category, \textit{i.e.}, \cite{BoykovJ01,ChengPZTR15,DongSSY15,FengPCC16,WangHC14,Grady06,GulshanRCBZ10,KassWT88,MortensenB95,RotherKB04,XianZCXD16,BadoualSUU17,XuPCYH16,XuPCYH17,LiewWXOF17,ManinisCTG18}, in which the user directly specifies the location of each label via seeds/scribbles \cite{BoykovJ01,DongSSY15,FengPCC16,Grady06,GulshanRCBZ10,WangHC14,XuPCYH16,LiewWXOF17,ManinisCTG18}, contours \cite{KassWT88,MortensenB95,XianZCXD16,BadoualSUU17,XuPCYH17}, or bounding boxes \cite{ChengPZTR15,RotherKB04,XuPCYH17}. These algorithms use {\em graph cuts}, {\em random walks}, {\em level set}, {\em geodesic distance}, or {\em deep network} to segment the images according to the user annotations.}

In general, different assignments of label locations often yield different segmentation results, which means that the user has the responsibility to specify good label locations for generating satisfactory segmentation results.
In contrast, our algorithm takes the responsibility to actively explore the informative image regions as the query seeds for the user.

\paragraph{Indirect interactive image segmentation.}
Another line is the indirect interactive image segmentation \cite{BatraKPLC10,FathiBRR11,KowdleCGC11,RupprechtPN15,StraehleKKBDH12}, in which the algorithms usually {\em recommend} several uncertain regions to the user, and then the segmentation algorithms adopt the user-selected regions for updating the segmentation results.
Batra~\textit{et al.} \cite{BatraKPLC10} propose a co-segmentation algorithm that provides the suggestion about where the user should draw scribbles next.
Based on the active learning method, Fathi~\textit{et al.} \cite{FathiBRR11} present an incremental self-training video segmentation method to ask the user to provide annotations for gradually labeling the frames.
For scene reconstruction, Kowdle~\textit{et al.} \cite{KowdleCGC11} also employ an active learning algorithm to query the user's scribbles about the uncertain regions.
To segment a large 3D dataset, Straehl~\textit{et al.} \cite{StraehleKKBDH12} provide various uncertainty measurements to suggest the user some candidate locations, and then segment the dataset using the watershed cut according to the user-selected locations.
Rupprecht~\textit{et al.} \cite{RupprechtPN15} model the segmentation uncertainty as a probability distribution over the set of sampled figure-ground segmentations, the collected segmentations are used to calculate the most uncertain region to ask the label from the user.
Chen~\textit{et al.} \cite{ChenCC16} select the query-pixel with the highest uncertainty referred to the transductive inference measurement.

This category of interactive image segmentation proposes the candidate label locations for the user, which eases the user's responsibilities of selecting good label locations for guiding the segmentation.
However, to provide the user with candidate label locations, the algorithms in this category usually take perceivable time to estimate the label locations and the user usually has to carefully label these locations in several clicks per round. In contrast, our seed proposal is very efficient and the user only has to effortlessly and unambiguously provide one swipe stroke per round.

\paragraph{Proposal generation.}
The purpose of object proposal generation \cite{ArbelaezPBMM14,CarreiraS10,ManenGG13,UijlingsSGS13,WangZLZJW15,XiaoLTLT15} is to provide a relatively small set of bounding boxes or segments covering probable object locations in an image, so that an object detector does not have to examine exhaustively all possible locations in a sliding window manner.
To increase the recall rate for object detection, a common solution in proposal generation is to diversify the proposals.
For example, Carreira and Sminchisescu \cite{CarreiraS10} present a diversifying strategy, which is based on the maximal marginal relevance measure \cite{CarbonellG98}, to improve the object detection recall.
Besides diversifying the proposals in spatial domain, diversifying the proposals by their similarities in feature domain has also been adopted \cite{ManenGG13,UijlingsSGS13,WangZLZJW15,XiaoLTLT15}.

In a similar manner, we diversify the selected query seeds in spatial and feature domains to improve the segmentation recall derived from the relatively small set of query seeds.

\section{Problem Statement}
Consider an image $\mathcal{I}$ that is represented as a graphical model over a vertex set $\mathcal{V} = \{ v_1, \cdots, v_{|\mathcal{V}|} \}$, where a vertex can be, for example, a pixel or a superpixel, or even an aggregation of neighboring superpixels. Assume that we have two kinds of labels $\mathsf{1}$ and $\mathsf{0}$, where $\mathsf{1}$ denotes the ROI and $\mathsf{0}$ denotes the non-ROI. If there exists a set of seeds $\mathcal{Q} = \{ q_1, \cdots, q_{|\mathcal{Q}|} \} \subseteq \mathcal{V}$ for user to label, then the user's labeling can be defined as a mapping $\eta: \mathcal{Q} \rightarrow \{\mathsf{0}, \mathsf{1}\}$, and hence the labeled-seed set can be defined as $\mathcal{U} = \{ \eta(q_1), \cdots, \eta(q_{|\mathcal{Q}|}) \}$. Based on the information \djc{from} the labeled-seed set $\mathcal{U}$, an interactive image segmentation algorithm aims to automatically partition the entire vertex set $\mathcal{V}$ as non-ROI vertex set $\mathcal{V}_\mathsf{0} \subseteq \mathcal{V}$ and ROI vertex set $\mathcal{V}_\mathsf{1} \subseteq \mathcal{V}$. In general, a segmentation algorithm may include a label propagation procedure, which can be defined as another mapping $\pi:\mathcal{V} \rightarrow [\mathsf{0}, \mathsf{1}]$ with $\mathcal{U}$ as its hints. We denote the segmentation generated via label propagation mapping $\pi$ as $\mathbf{s}^\mathbf{\pi}$, and all possible segments of $\mathcal{V}$ as the set $\mathcal{S} =\{\mathsf{0}, \mathsf{1}\}^{|\mathcal{V}|}$. 

\subsection{Interactive Image Segmentation}
The general interactive image segmentation problem can be formulated as follows: Given an image $\mathcal{I}$ and a label propagation algorithm $\pi$, the user specifies a labeled-seed set $\mathcal{U}$ to make the machined-generated segmentation $\mathbf{s}^\pi$ approach the expected segmentation $\mathbf{s}^*$. 

We use a conditional probability $p( \mathbf{s}^\pi = \mathbf{s}^* |\mathcal{U})$ over $\mathcal{S}$ to state how likely it can be for a segmentation $\mathbf{s}^\pi$ to approximate $\mathbf{s}^*$. For brevity, we denote the probability as $p( \mathbf{s}^\pi |\mathcal{U})$ in the rest of the paper. We would like to model the distribution over $\mathcal{S}$, but it is intractable to do exhaustive computation of $p( \mathbf{s}^\pi |\mathcal{U})$ over $\mathcal{S}$, since the cardinality of $\mathcal{S}$ is extremely large ($2^{|\mathcal{V}|}$). There are basically two strategies to make $\mathbf{s}^\pi$ approach $\mathbf{s}^*$ and hence to maximize the conditional probability $p( \mathbf{s}^\pi |\mathcal{U})$. The first one is done by improving the performance of label propagation algorithm $\pi$ and the second one is to improve the quality of the seeds $\mathcal{Q}$. Improving label propagation may help to achieve better label inference for unlabeled vertices. \djc{Many interactive image segmentation algorithms have explored this direction~\cite{BoykovJ01,ChengPZTR15,DongSSY15,FengPCC16,WangHC14,Grady06,GulshanRCBZ10,KassWT88,MortensenB95,RotherKB04,XianZCXD16,BadoualSUU17,XuPCYH16,XuPCYH17,LiewWXOF17,ManinisCTG18}. With the aids of deep networks, the deep learning based algorithms \cite{XuPCYH16,XuPCYH17,LiewWXOF17,ManinisCTG18} especially show good performance in this direction. However, the issue of seed-selection is left to the user.} Our approach, on the other hand, focuses on how to select the informative query seeds $\mathcal{Q}$ for the user to annotate and therefore eases the annotation burden.

\subsection{Diversified Seed Proposals}
Given a label propagation algorithm $\pi$, in order to make a segmentation $\mathbf{s}^\pi$ approach the expected $\mathbf{s}^*$ in fewer rounds, we propose to select the \emph{informative query seeds} $\mathcal{Q}$ under the criterion of maximizing  the improvement on conditional probability $p(\mathbf{s}^\pi|\mathcal{U})$.

Our idea of selecting new query seeds $\mathcal{Q}'$ at each round of interaction can be formulated as the following optimization problem:
\begin{equation}\label{eq:funcProblem}
  \begin{aligned}
  \underset{\footnotesize \begin{array}{c} \mathcal{Q}' \subseteq \mathcal{V} \\ |\mathcal{Q}'|=\theta_k\end{array}}{\text{max}}
  & \sum_{v_i \in \mathcal{Q}'} p(\mathbf{s}^\pi | \widetilde{\mathcal{U}} \cup \eta(v_i) ) - p(\mathbf{s}^\pi | \widetilde{\mathcal{U}}) \\
  \text{s.t.} \quad
  & d(v_i, v_j) \geq \theta_d, \, \forall v_i,v_j \in \mathcal{Q}' \,,
  \end{aligned}
\end{equation}
where $\theta_k$ denotes the number of query seeds being selected at each round, $\eta(v_i)$ defines the label of $v_i$, $\widetilde{\mathcal{U}}$ denotes all labels obtained in the previous rounds, $d(\cdot, \cdot)$ computes the Euclidean distance, and $\theta_d$ denotes the spatial distance on the touchscreen. The objective function in Eq.~(\ref{eq:funcProblem}) aims to find a new query set $\mathcal{Q}'$ of size $\theta_k$ that yields the maximal improvement. The distance constraint is to guarantee that the query seeds are separated far enough from each other on the touchscreen so that the user is able to swipe through the seeds effortlessly and unambiguously.

Expanding the set of labeled seeds $\widetilde{\mathcal{U}}$ is always helpful for approximating the expected segmentation $\mathbf{s}^*$ by $\mathbf{s}^\pi$ because more hints can be obtained from the user. However, the difficulty in optimizing Eq.~(\ref{eq:funcProblem}) is how to select the most informative query seeds that increase the probability most. Since the expected segmentation $\mathbf{s}^*$ is not given, it is hard to evaluate the contribution of each query seed. We tackle the problem through an observation: If the results of two label propagations are similar, their conditional probabilities should also be similar and thus do not yield significant improvements. Therefore, we propose to select the vertices that have higher chance to produce greater change in label propagation as the informative query seeds. The selection is carried out via our query-seed proposal scheme, which is described later in Section \ref{sec:seedProposal}. 

In \djc{summary}, to make the estimated segmentation $\mathbf{s}^\pi$ approach the expected segmentation $\mathbf{s}^*$ more quickly in fewer interactions, we select and propose the informative query seeds $\mathcal{Q}'$ to the user for acquiring reliable labels that can be used to guide the label propagation algorithm $\pi$. Furthermore, Eq.~(\ref{eq:funcProblem}) is designed not only for proposing the informative query seeds $\mathcal{Q}'$ but also for making sure that the new query seeds $\mathcal{Q}'$ are easy to label via swipe gestures.

\section{Approach}
An intuitive description of our interactive segmentation approach is as follows. We represent the input image as a weighted two-layer graph. At each round of user-machine interaction, we propose $\theta_k$ query seeds to acquire the true labels from the user. Then, the labeled vertices propagate their labels to the rest unlabeled vertices, and thus yield the corresponding segmentation. According to the clues of label propagation, the other $\theta_k$ query seeds are then proposed to the user for the next interaction. In the experiments, the user-machine interaction is repeated until we have performed a predefined number of rounds. An overview of our approach is shown in Fig.~\ref{fig:overview}.

As previously mentioned, we use the conditional probability $p(\mathbf{s}^\pi | \mathcal{U})$ to state how likely the estimated segmentation $\mathbf{s}^\pi$ is equal to the expected segmentation. We use an EM-like procedure to maximize the conditional probability by alternately performing {\em i}) a query-seed proposal scheme to find $\mathcal{Q}'$ with respect to Eq.~(\ref{eq:funcProblem}) and {\em ii}) a label propagation scheme to find $\mathbf{s}^\pi$ guided by $\mathcal{U}$.

\subsection{Query-seed Proposal Scheme} \label{sec:seedProposal}
Selecting query seeds is based on \emph{seed assessment} and \emph{seed diversification}, which jointly find approximate solutions to Eq.~(\ref{eq:funcProblem}).
The step of seed assessment ranks the query seeds according to \emph{proposal confidence} and \emph{proposal influence}.
The step of seed diversification is to satisfy the constraint in Eq.~(\ref{eq:funcProblem}).

We denote the labeled vertex set and the unlabeled vertex set as $\mathcal{V}_L$ and $\overline{\mathcal{V}_L}$, respectively. The query-seed proposal scheme extracts a $\theta_k$-element subset $\mathcal{Q}' \subseteq \overline{\mathcal{V}_L}$ per round to acquire their true labels from the user. After the labels of $\mathcal{Q}'$ are acquired, we merge $\mathcal{Q}'$ into $\mathcal{V}_L$.

\subsubsection{Seed Assessment} \label{sec:assessment}
While building the query seed set $\mathcal{Q}'$, we use an assessing function $f:\mathcal{V} \rightarrow \mathbb{R}$ to assign each unlabeled seed $v_i$ a value $f(v_i)$ accounting for \djc{all labeled vertices in the previously labeled set $\widetilde{\mathcal{Q}}$, which is the union of two disjoint sets $\widetilde{\mathcal{Q}}_\mathsf{0}$ and  $\widetilde{\mathcal{Q}}_\mathsf{1}$ according to the types of labels. The function $f$ has the following form:
    \begin{equation}\label{eq:funcScore}
        f(v_i|\widetilde{\mathcal{Q}}) = \Phi(v_i|\widetilde{\mathcal{Q}}) + \theta_s \Psi(v_i) \,,
    \end{equation}
where $\widetilde{\mathcal{Q}}=\widetilde{\mathcal{Q}}_\mathsf{0} \bigcup \widetilde{\mathcal{Q}}_\mathsf{1}$, $\theta_s$ is a weighting factor. We use $\theta_s = 0.7$ in the experiments.}

The first term $\Phi$ in Eq.~(\ref{eq:funcScore}) computes the proposal confidence. Let $\rho:\mathcal{V} \times \mathcal{V} \rightarrow \mathbb{R}$ denote a metric that can estimate the graph distance\footnote{Note that the graph distance here is weighted with respect to the adopted features, not merely defined in the spatial domain.} of each unlabeled vertex to any specified vertex. The vertices within a short graph distance have high chance to share the same label, and thus are more likely to be redundant queries. Hence, a vertex that is distant from the other labeled vertices should be more informative and suitable to be selected as a query for acquiring label. Hence we let the proposal confidence of a vertex be proportional to its graph distance to the nearest labeled vertex. 

The second term $\Psi$ in Eq.~(\ref{eq:funcScore}) calculates the proposal influence. We use this term to define the influence of a vertex. This term is inspired by semi-supervised learning \cite{ChapelleWS02,ZhouBLWS03} with the assumption of \emph{label consistency}. Label consistency means the vertices on the same manifold structure or nearby vertices are likely to have the same label. A vertex has more similar vertices around it should has larger influence, since there might be more similar vertices sharing the same label with it. We let the proposal influence of a vertex be proportional to the number of similar neighboring vertices around it.


Any graph distance measurement and clustering algorithm can be used to estimate the graph distance and to extract the similar neighboring vertices. 
We choose to use the shortest path to estimate the graph distance for the proposal confidence term, and use a minimum spanning tree algorithm to extract similar neighboring vertices for the proposal influence term. 
The two algorithms are chosen for their computational efficiency. Section \ref{sec:implementation} details the implementation of the two terms.

\subsubsection{Seed Diversification}
\label{sec:diversification}
Since we would like to acquire $\theta_k$ true labels from the user via swipe gestures per interaction, the multiple query seeds should be sufficiently distant from one another, as modeled in the constraint of Eq.~(\ref{eq:funcProblem}), so that the user is able to swipe through the seeds effortlessly and unambiguously.

The seed diversification step sorts the query vertices from high assessment values to low assessment values and then performs non-maximum suppression: If a vertex within the radius of $\theta_d$ pixels to any already selected higher-valued vertex, we just skip this vertex and move on to the next one until we get totally $\theta_k$ vertices as the query seeds. Note that the skipped vertices may be reconsidered in the subsequent rounds. 

\subsection{Label Propagation Scheme}\label{sec:labelProp}
Label propagation is used to propagate the known labels to all other not-queried vertices.
A result of segmentation can be obtained by directly assigning each vertex the same label as its closest vertex that has already been labeled by the user.
Here we also use the shortest path on graph as in seed assessment to compute the closeness between vertices for label propagation.

\subsection{Implementation Details}\label{sec:implementation}
\subsubsection{Graph Construction}
We design a two-layer weighted graph $\mathcal{G}=(\mathcal{V},\mathcal{E},\omega)$ for selecting the query seeds and generating the segmentation. The graph consists of moderate-granularity vertices (superpixel-level) and large-granularity vertices (tree-level). The tree-level vertices are used to \emph{guide the superpixel-level vertices} to perform query-seed assessment and label propagation. The use of tree-level vertices is just to entail the aforementioned assumption of \emph{label consistency}, and we do not consider the tree-level vertices as query seeds.

\textbf{Vertices.}
We first over-segment an input image into a set of superpixels $\mathcal{R} = \{ r_1, r_2, \cdots, r_{|\mathcal{R}|} \}$ using the SLIC algorithm \cite{AchantaSSLFS12}. The set $\mathcal{R}$ is then partitioned into a minimum-spanning-tree (MST) set $\mathcal{T} = \{ t_1, t_2, \cdots, t_{|\mathcal{T}|} \}$ using the Felzenszwalb-Huttenlocher (FH) algorithm \cite{FelzenszwalbH04}. For each tree $t_i$ in the FH algorithm, the $\tau$ function is used as a threshold function for merging superpixels and is defined as
\begin{equation}\label{eq:tau}
    \tau(t_i) = \frac{\theta_t}{|t_i|} \,,
\end{equation}
where $|t_i|$ is the size of $t_i$ in pixels, and $\theta_t$ controls the number of trees. \djc{In the FH algorithm, two spatially adjacent MSTs are merged if the feature difference between them is smaller than the respective internal feature difference. Since the value of $\tau$ is embedded as a fundamental internal feature difference of each MST, a larger value $\theta_t$ causes higher internal feature difference and thus encourages the merging. Therefore, a larger value $\theta_t$ means fewer yet larger trees will be generated.} 

\djc{The minimum-spanning-tree algorithm is used to construct the tree-level vertices, in which each tree-level vertex is associated with some superpixel-level vertices. Fig.~\ref{fig:overview} shows a schematic diagram of the vertex sets $\mathcal{R}$ and $\mathcal{T}$. In our approach, the tree-level vertices provide shortcuts between superpixel-level vertices and information for calculating the proposal influence.}

Given the vertex set $\mathcal{V} = \{ \mathcal{R} \cup \mathcal{T} \}$, we have to compute the features of each vertex. We use the normalized color histograms $h_c$ with 25 bins for each CIE-Lab color channel. We also include the texture feature consisting of Gaussian derivatives in eight orientations, which are quantized by magnitude to form a normalized texture histogram $h_t$ with ten bins for each orientation per color channel. Hence, each vertex $v_i \in \mathcal{V}$ is represented by the histograms $h_c$ and $h_t$.

\textbf{Edges.}
The edge set $\mathcal{E}$ is defined according to the vertex types. An edge $e_{ij} \in \mathcal{E}$ exists if 1) two vertices $r_i,r_j \in \mathcal{R}$ are adjacent, 2) two vertices $t_i,t_j \in \mathcal{T}$ are adjacent, or 3) vertex $r_i \in \mathcal{R}$ is included in its corresponding tree-level vertex $t_j \in \mathcal{T}$. \djc{Here, by `adjacent' we mean two vertices are adjacent spatially. Please refer to the green lines of the `Graph Construction' in Fig.~\ref{fig:overview}.} 

Given two vertices $v_i, v_j \in \{\mathcal{R} \cup \mathcal{T}\}$, we use the following equation \cite{GrundmannKHE10a} to measure the inter-vertex feature distance:
\begin{equation} \label{eq:colorTexture}
    \omega(v_i,v_j) = \bigg (\, 1-\left(1-\chi^2_{h_c}(v_i,v_j) \right) \left(1-\chi^2_{h_t}(v_i,v_j) \right) \, \bigg )^2~,
\end{equation}
where $\chi^2_{h_c}(v_i,v_j)$ is the $\chi^2$ color distance, $\chi^2_{h_t}(v_i,v_j)$ is the $\chi^2$ texture distance. The equation makes the inter-vertex feature distance close to zero only if both the color and texture distances are close to zero.

Having the two-layer weighted graph $\mathcal{G}=(\mathcal{R} \cup \mathcal{T},\mathcal{E},\omega)$ been defined, we can now estimate the similarities between vertices and labeled seeds using the notion of shortest path on graph \cite{GulshanRCBZ10,KrahenbuhlK14,RupprechtPN15,WangSP15,WeiWZS12}. 

\subsubsection{Graph Distance}
Given the two-layer graph $\mathcal{G}=(\mathcal{V},\mathcal{E},\omega)$ with a specified vertex $v_j \in \mathcal{V} = \{\mathcal{R} \cup \mathcal{T}\}$, the geodesic distance $\Phi(v_i|v_j)$ of the shortest path from vertex $v_i$ to the specified vertex $v_j$ is defined as the accumulated edge weights along the path.
The geodesic distance function $\Phi$ can be defined as
\begin{equation} \label{eq:geoDist}
\Phi(v_i|v_j) = \min_{v'_1=v_i,\ldots,v'_m=v_j} \sum_{k=1}^{m-1} \omega(v'_k,v'_{k+1}), \forall v'_k,v'_{k+1} \in \mathcal{V} \,,
\end{equation}
where $m$ denotes the path length.

\subsubsection{Proposal Confidence and Segmentation}
According to the geodesic distance function $\Phi$ in Eq.~(\ref{eq:geoDist}), a shorter distance between two vertices means they have higher label similarity, and we use the $\Phi$ function to derive seed assessment in Eq.~(\ref{eq:funcScore}) and label propagation in Section \ref{sec:labelProp}, \textit{i.e.}, the mapping function $\pi$ from $\mathcal{V}$ to $[0,1]$.

\subsubsection{Proposal Influence}
The influence of each vertex $v_i$ is defined as
\djc{
\begin{equation}\label{eq:szTree}
    \Psi(v_i) = \frac{ \max \big\{ |t_j| \, \big\rvert v_i \in t_j \big\} }{| \mathcal{I} |}, \forall v_i \in \mathcal{V}, t_j \in \mathcal{T} \,,
\end{equation}
}
where the function $|\cdot|$ extracts the size of vertex in pixels. In the numerator, the larger size means a superpixel-level vertex $v_i$ is included in the corresponding tree-level vertex $t_j$ together with more similar superpixel-level vertices, which means $v_i$ has higher influence power. 

\subsubsection{The Seed Assessment Function}
By plugging Eq.~(\ref{eq:geoDist}) and Eq.~(\ref{eq:szTree}) into Eq.~(\ref{eq:funcScore}), we can define the assessment function for each vertex $v_i$ with respect to \djc{the previously labeled vertex set $\widetilde{\mathcal{Q}}$:
\begin{equation}\label{eq:obj}
    f(v_i|\widetilde{\mathcal{Q}}) = \Phi(v_i|\widetilde{\mathcal{Q}}) + \theta_s  \Psi(v_i), \; \forall v_i \in \mathcal{R} \,.
\end{equation} }
The seed assessment criterion of Eq.~(\ref{eq:obj}) and the seed diversification constraint described in Section \ref{sec:diversification} are used for choosing the \djc{superpixel-level} vertices as the query seeds that help to solve the optimization in Eq.~(\ref{eq:funcProblem}).
Each selected \djc{superpixel-level} vertex should have a distinct feature and belong to a tree of larger size, and should be at a larger spatial distance to the previously labeled vertex set $\widetilde{\mathcal{Q}}$.

Notice that, we always use the centroid pixel of the selected vertex to represent the query seed on the display. The first query in our algorithm is the centroid superpixel of the largest tree since the labeled vertex set is empty. The selection of subsequent seeds then follows the rule of Eq.~(\ref{eq:obj}).

\section{Experimental Results}
We conduct four kinds of experiments to evaluate our approach in depth.
The first experiment compares the segmentation accuracy of different parameter settings in our approach.
The second and the third experiments compare our approach with the state-of-the-art algorithms in terms of segmentation accuracy and computation time, where the scenario of interaction could be one seed per interaction or multiple seeds per interaction.
The fourth experiment provides the user study.
More experimental results can be found in the supplementary material.


\textbf{Datasets.}
Fig.~\ref{fig:datasetGt} illustrates some examples of the ground-truth segments of the six datasets used in our experiments.
\begin{enumerate}[]
\item \emph{SBD} \cite{GouldFK09}: This dataset contains 715 natural images. Each image has average $4.22$ ground-truth segments in each individual annotation.
\item \emph{ECSSD} \cite{ShiYXJ16}: This dataset contains 1000 natural images. Each image has average $1.0$ ground-truth segment in each individual annotation.
\item \emph{MSRA}\footnote{The individual annotation are provided by Achanta~\textit{et~al.} \cite{AchantaHES09}. The natural images are provided by Liu~\textit{et al.} \cite{LiuSZTS07}.} \cite{AchantaHES09}: This dataset contains 1000 natural images. Each image has average $1.0$ ground-truth segment in each individual annotation.
\item \emph{VOC} \cite{VOC07}: We use the {\em trainval} segmentation set, which contains 422 images. Each image has average $2.87$ ground-truth segments in each individual annotation.
\item \emph{BSDS} \cite{FowlkesMM07}:
    This dataset contains 300 natural images.
    Each image has several hand-labeled segmentations as the ground truths. Each image has average $20.37$ ground-truth segments in each individual annotation.
\item \emph{IBSR}\footnote{The MR brain data sets and their manual segmentations were provided by the Center for Morphometric Analysis at Massachusetts General Hospital and are available at http://www.cma.mgh.harvard.edu/ibsr/.}: There are 18 subjects in this dataset. For each subject we extract 90 brain slices ranging from 20th slice to 109th slice.
\end{enumerate}

\begin{figure}[t]
    \centering
    \subfigure[SBD]  { \includegraphics[height=0.09\textwidth]{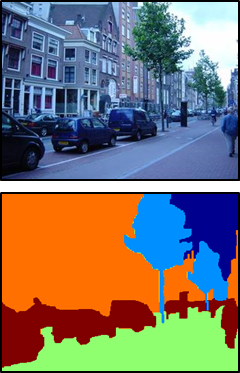} }
    \subfigure[{\tiny ECSSD}]{ \includegraphics[height=0.09\textwidth]{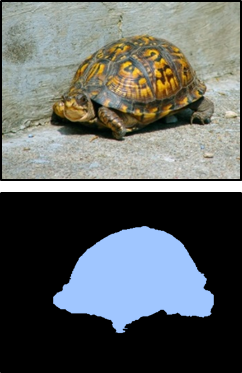} }
    \subfigure[MSRA] { \includegraphics[height=0.09\textwidth]{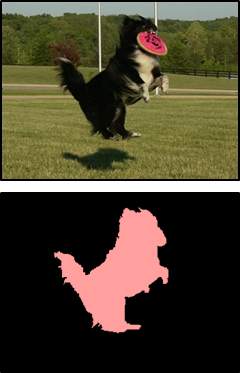} }
    \subfigure[VOC]  { \includegraphics[height=0.09\textwidth]{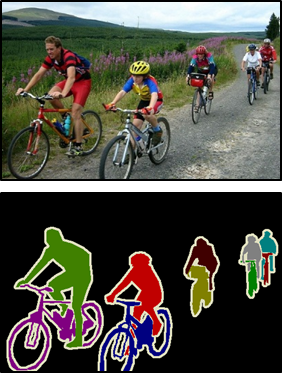} }
    \subfigure[BSDS] { \includegraphics[height=0.09\textwidth]{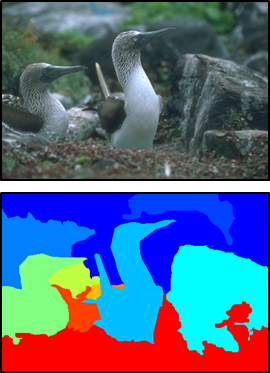} }
    \subfigure[IBSR] { \includegraphics[height=0.09\textwidth]{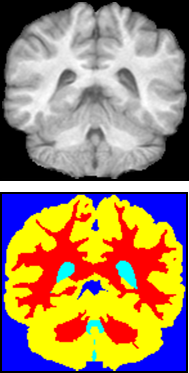} }
    \vspace{-3mm}
    \caption{\label{fig:datasetGt}  Examples of the ground-truth segments from each dataset. Each color denotes a ground-truth ROI except the black color in (b-d) and the creamy-white color in (d). Note that we only show one human annotation in (e) for better visualization.}
\end{figure}

\begin{figure*}[t]
    \centering
    \subfigure[VOC-$|\mathcal{R}|$]
        { \includegraphics[width=0.23\textwidth]{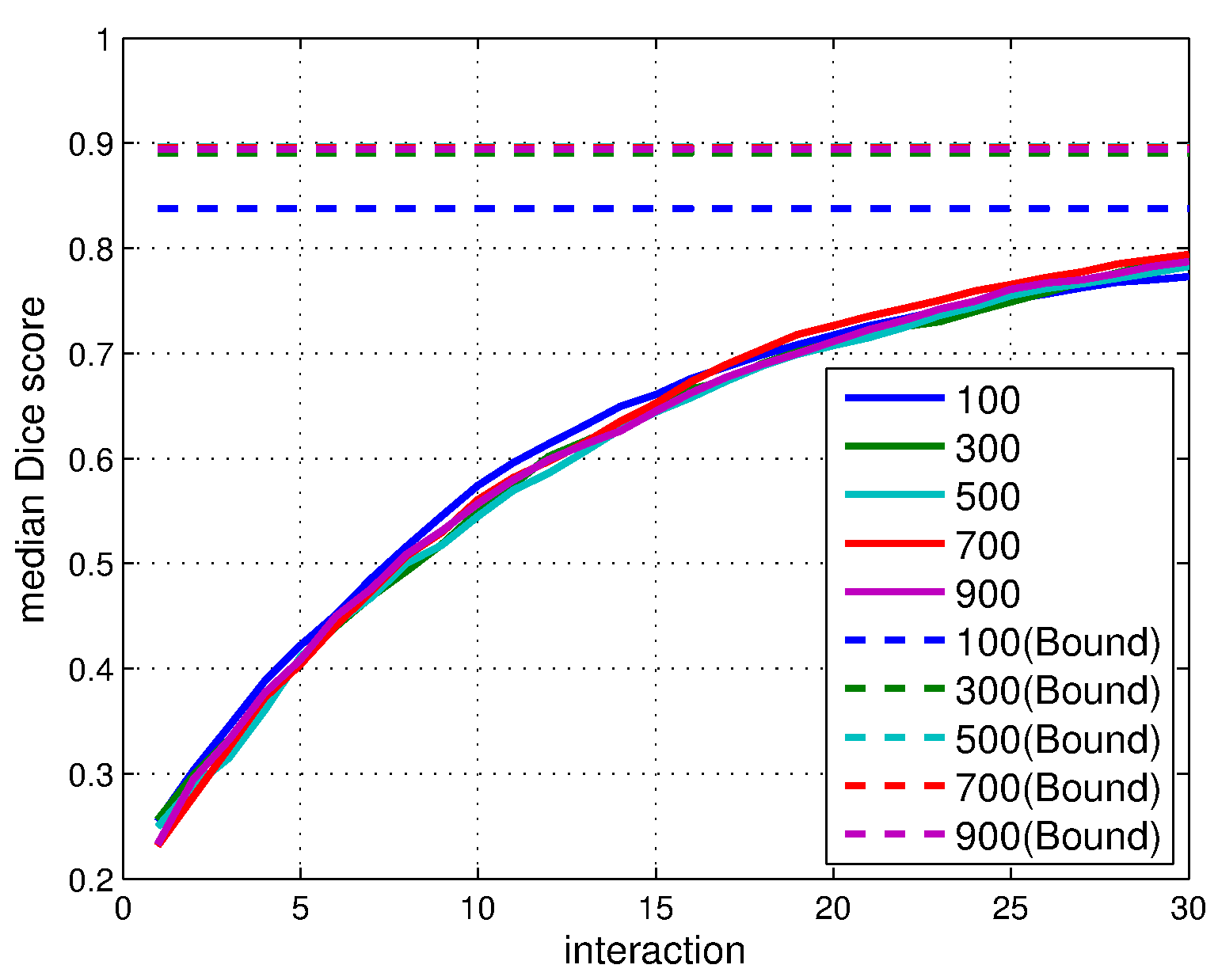} }
    \subfigure[VOC-$\theta_t$]
        { \includegraphics[width=0.23\textwidth]{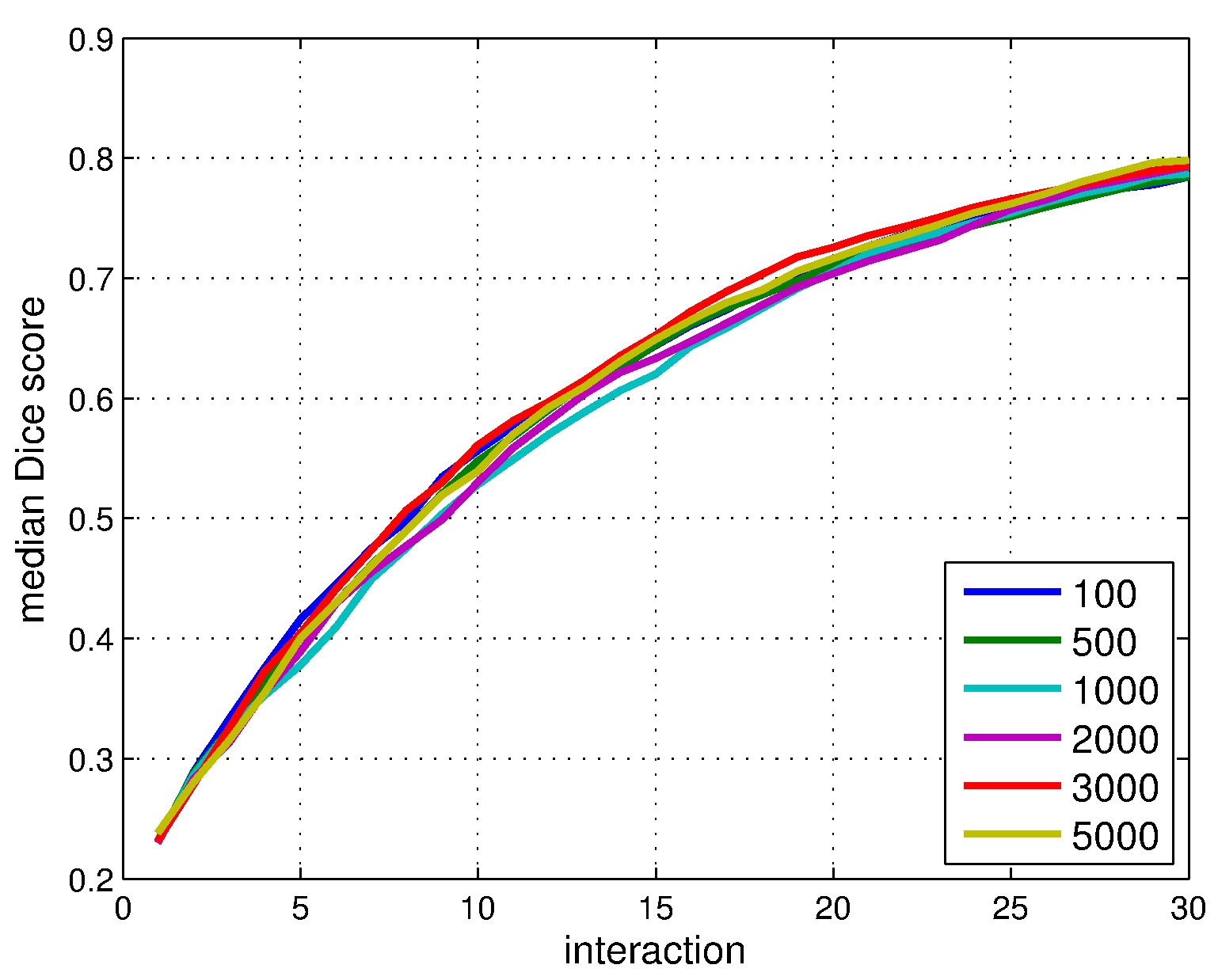} }
    \subfigure[VOC-$|t|$]
        { \includegraphics[width=0.23\textwidth]{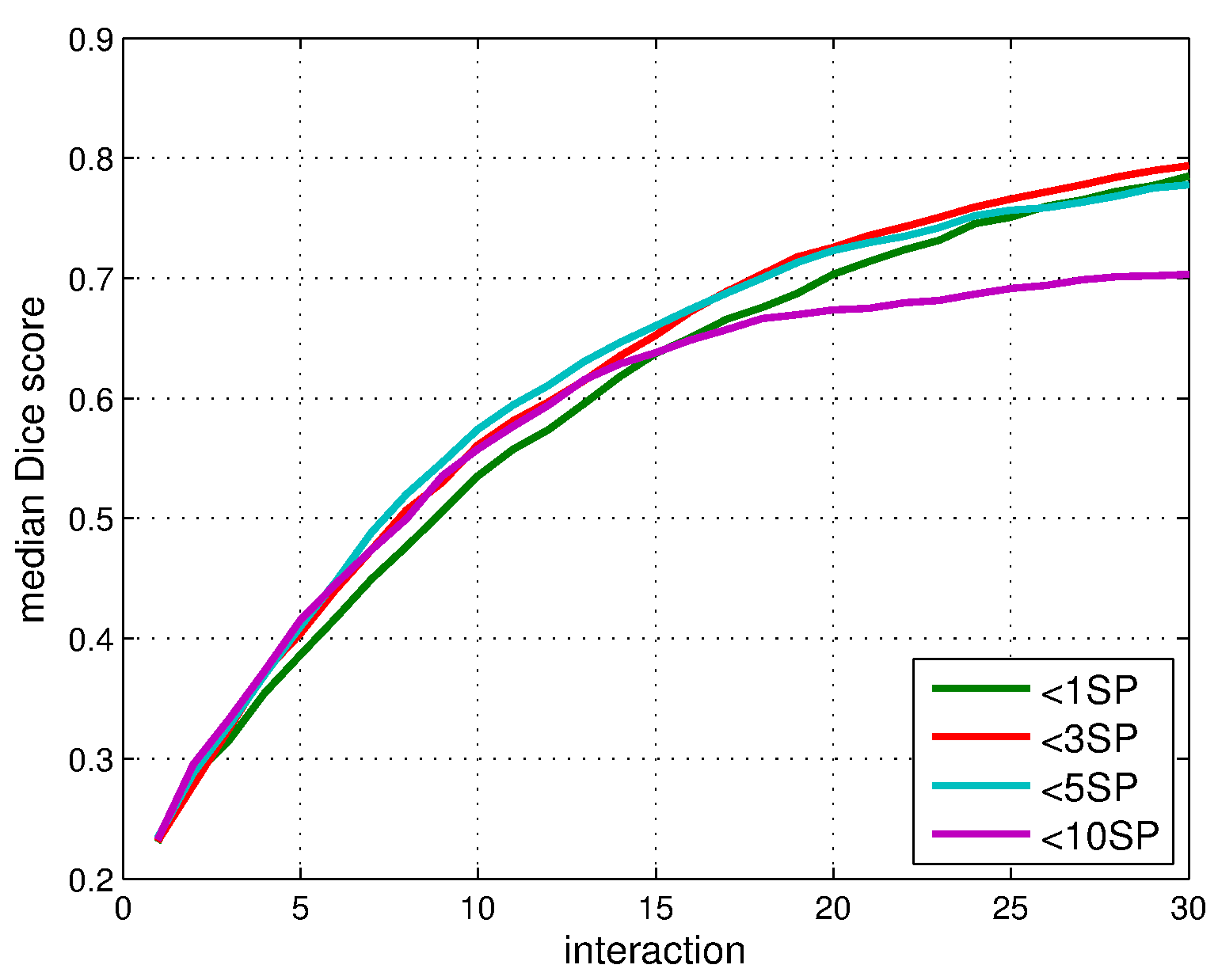} }
    \subfigure[VOC-$\theta_s$]
        { \includegraphics[width=0.23\textwidth]{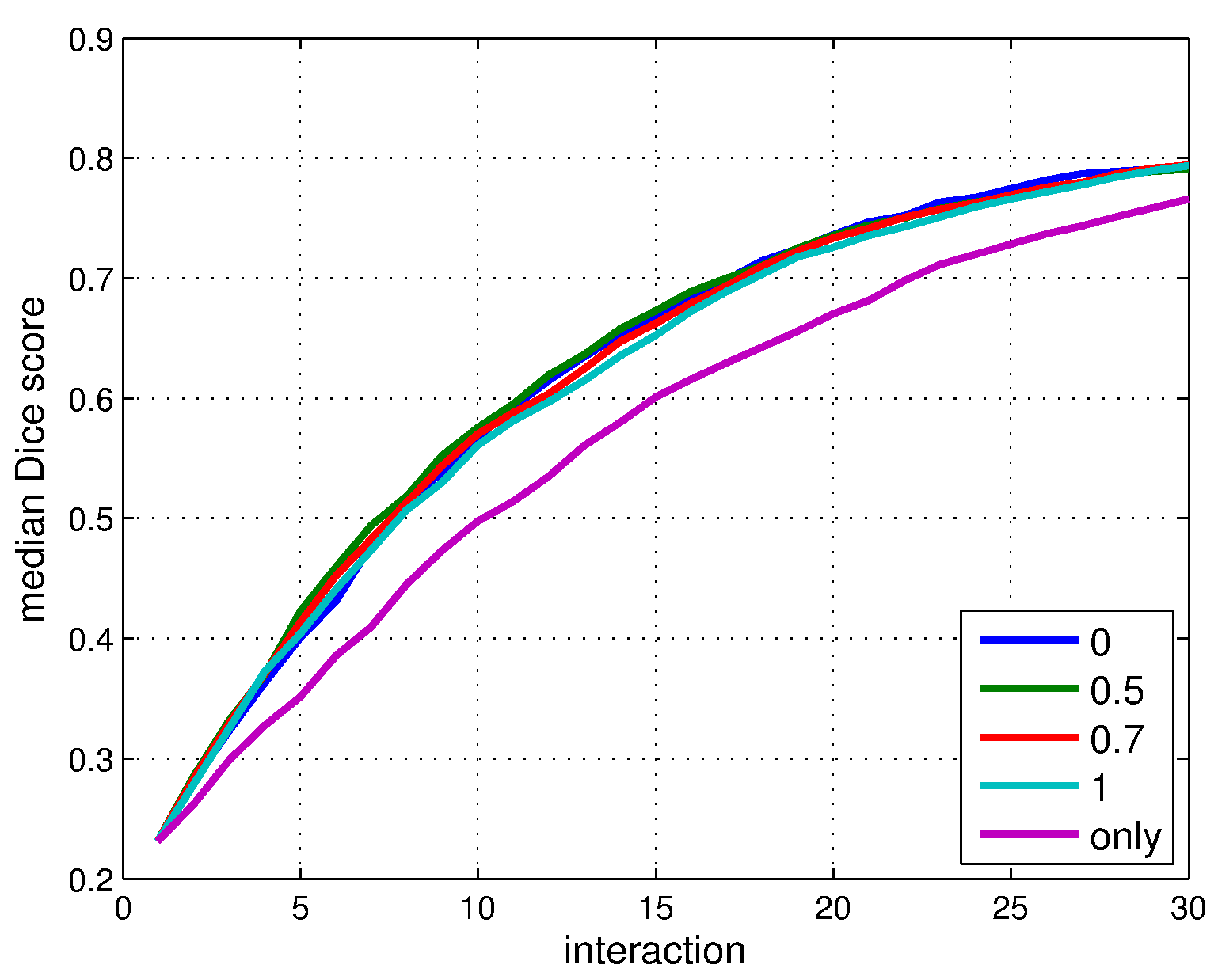} }
    \vspace{-3mm}
    \caption{\label{fig:comparison1} Comparison results on various parameter settings of our approach. Each sub-figure depicts the median Dice score as the segmentation accuracy against the number of interactions. The complete results are shown in the supplementary material.}
\end{figure*}

\textbf{Evaluation metric.}
For evaluating the segmentation accuracy, every segment in each individual annotation is considered as an ROI. For each ROI, we perform 30 rounds of interactive segmentation. The evaluation metric used to measure the segmentation quality is the median of the Dice score\footnote{For fair comparison, we evaluate our approach with this metric as \cite{ChenCC16,RupprechtPN15}.}. The Dice score \cite{Soensen48} is defined as
\begin{equation}\label{eq:funcDice}
    \mathrm{dice}(C,G) = \frac{2|C \bigcap G|}{|C|+|G|} \,,
\end{equation}
where $C$ denotes the computer-generated segmentation and $G$ denotes the ground-truth segmentation.

\subsection{Effects of Different Parameter Settings}
We compare four different parameter settings on MSRA datasets to explore the properties of the proposed algorithm. The performance is evaluated by the segmentation accuracy against the number of interactions.

Fig.~\ref{fig:comparison1}a shows the comparison results of choosing different settings on the number of superpixels during graph construction. For reference, we plot the optimal segmentation accuracy that can be achieved by different settings in dash lines. The optimal segmentation accuracy is obtained by assigning all superpixels the `correct' labels, which is equivalent to performing infinite rounds of interactions. Based on this experiment, we choose to use 700 superpixels for the subsequent experiments. Fig.~\ref{fig:comparison1}b and Fig.~\ref{fig:comparison1}c compare the different settings of building the minimum spanning trees. Setting a larger value of $\theta_t$ would favor constructing larger trees. The value of $|t|$ means the minimum size constraint of each tree, which is used to merge the trees smaller than a certain degree of average-superpixel-size to their adjacent trees.  Based on the experimental result, we set $\theta_t=3{,}000$ and require that the minimum tree must contain at least three superpixels. Fig.~\ref{fig:comparison1}d  compares different settings of the weighting factor in the seed assessment function \ref{eq:funcScore}. The legend `only' in Fig.~\ref{fig:comparison1}d means the seed assessment function contains only the proposal influence term. We set $\theta_s=0.7$.
Fig.~\ref{fig:comparison1} demonstrates that our approach is not sensitive to the parameter setting.

\begin{figure*}[t]
    \centering
    \subfigure[{\tiny VOC - one seed per interaction}]  { \includegraphics[width=0.31\textwidth]{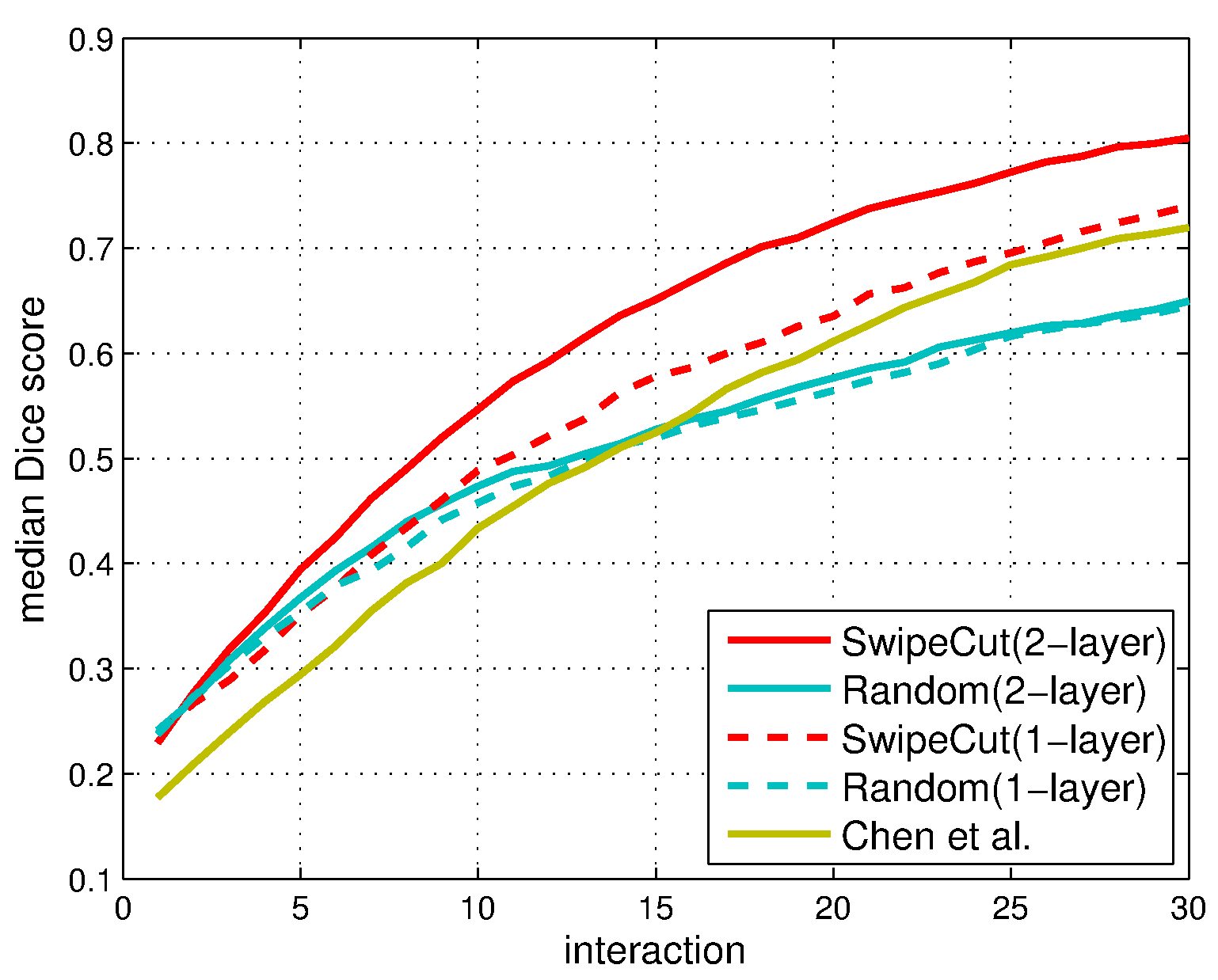} }
    \subfigure[{\tiny BSDS - one seed per interaction}] { \includegraphics[width=0.31\textwidth]{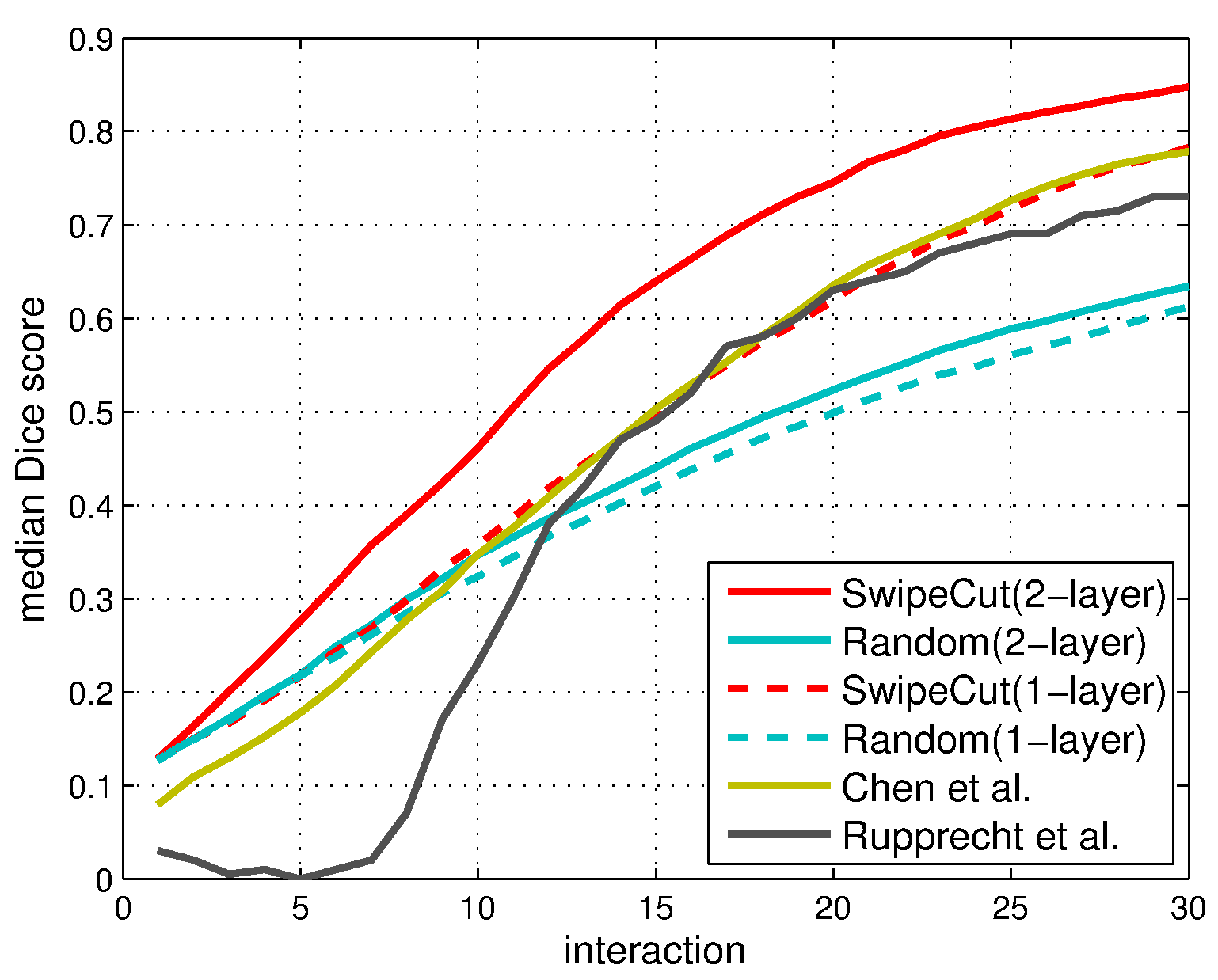} }
    \subfigure[{\tiny IBSR - one seed per interaction}] { \includegraphics[width=0.31\textwidth]{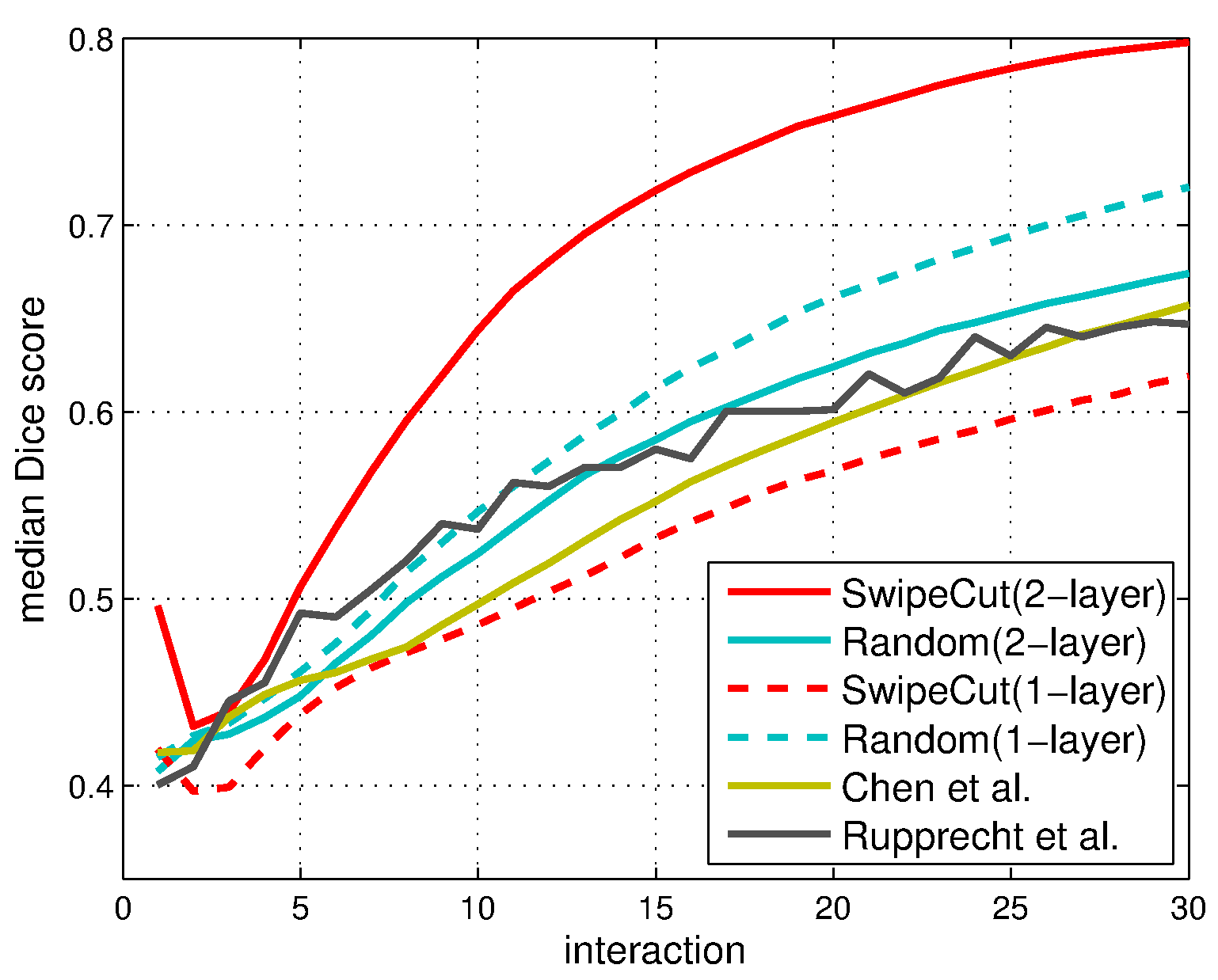} }
    \vspace{-3mm}
    \caption{\label{fig:comparison2} Comparison results on the state-of-the-art binary-query interactive segmentation algorithms \cite{ChenCC16,RupprechtPN15}. This figure shows the comparison of the segmentation accuracy. Each sub-figure depicts the median Dice score against the number of interactions. The results of Rupprecht~\textit{et al.} are reproduced according to \cite{RupprechtPN15}. Note that all methods in comparison propose seeds without the ground truth.}
\end{figure*}

\begin{figure*}[t]
    \centering
    \subfigure[{\tiny SBD - multi-seed per interaction}]  { \includegraphics[width=0.31\textwidth]{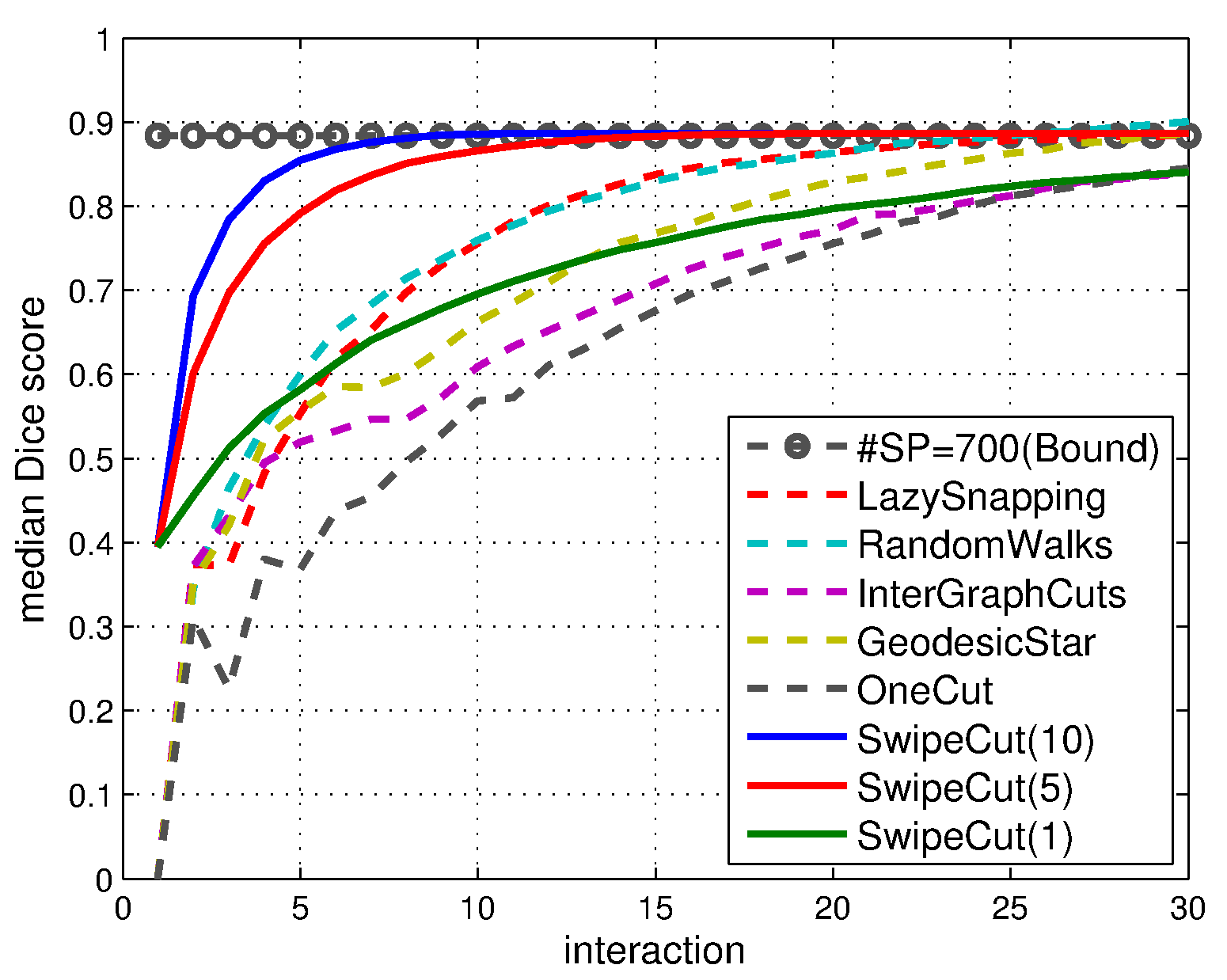} }
    \subfigure[{\tiny ECSSD - multi-seed per interaction}]{ \includegraphics[width=0.31\textwidth]{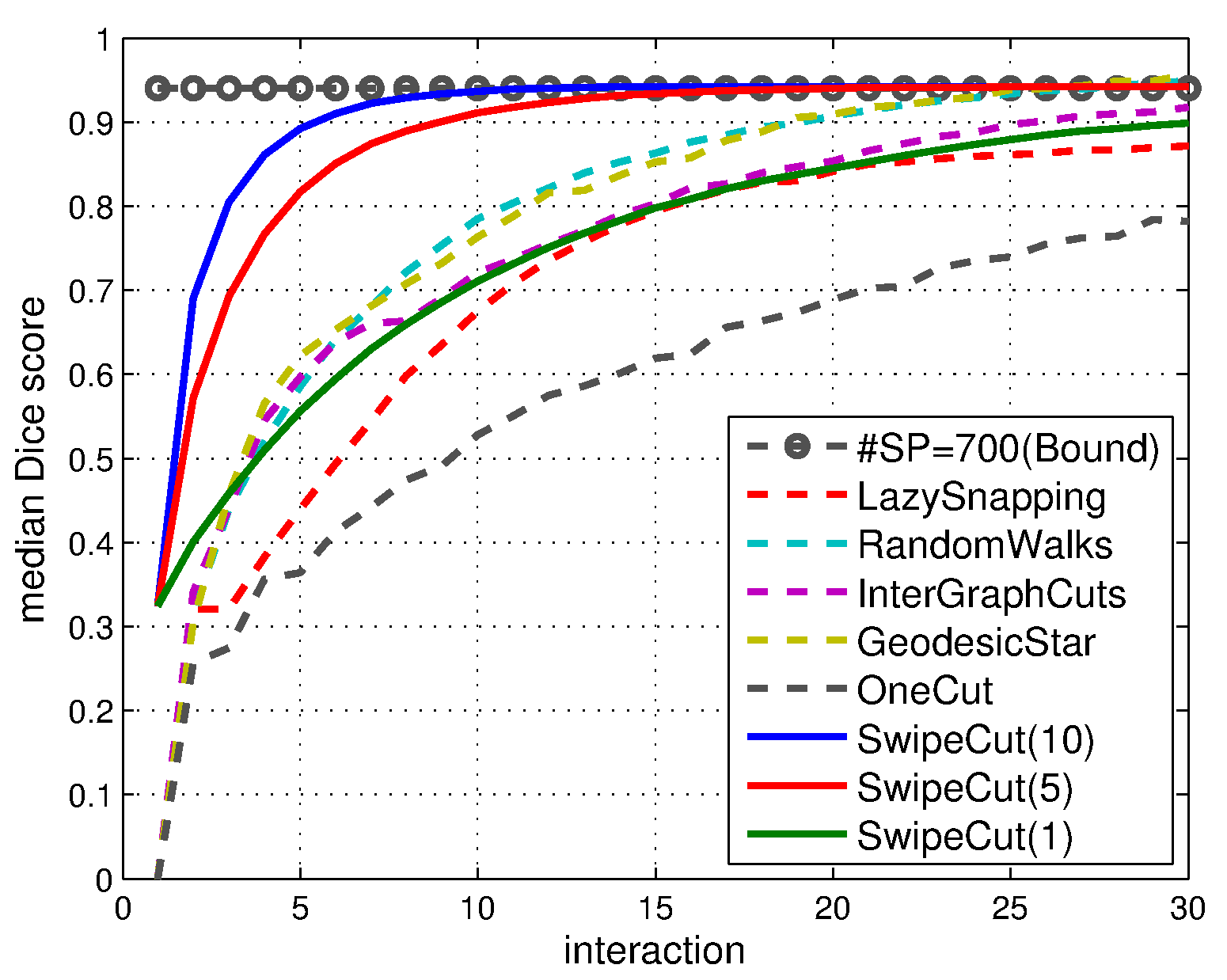} }
    \subfigure[{\tiny MSRA - multi-seed per interaction}] { \includegraphics[width=0.31\textwidth]{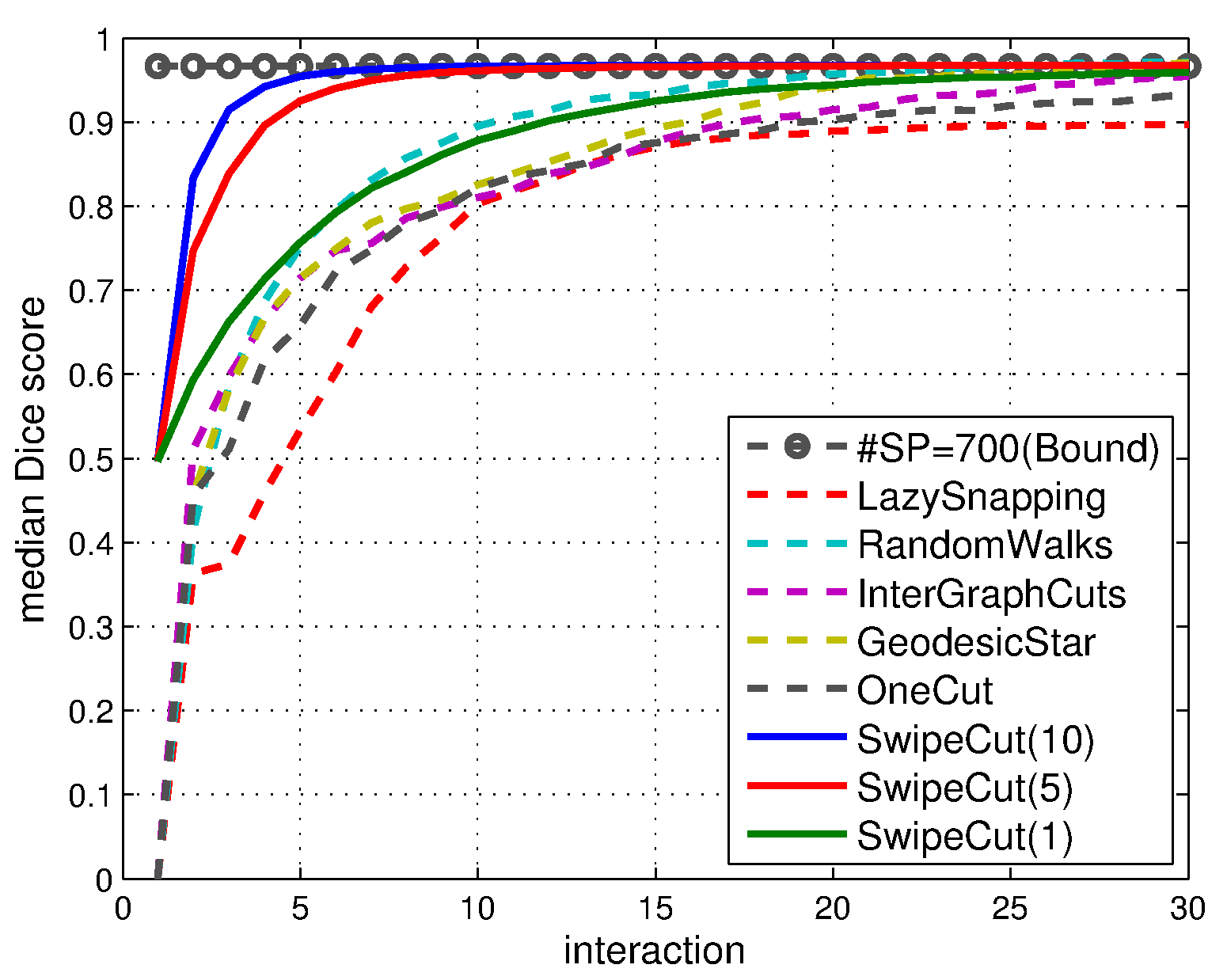} }
    \vspace{-3mm}
    \caption{\label{fig:comparison3} Comparisons on the variants of our approach and state-of-the-art methods. The performance is evaluated by the segmentation accuracy with respect to the number of interactions on six datasets. Each sub-figure depicts the median Dice score against the number of interactions. Note that all methods except the SwipeCut are annotated with the ground truth.}
\end{figure*}

\subsection{One Query Seed Per Interaction}
If we select only one seed per interaction, \textit{i.e.}, $\theta_k=1$, our approach is actually similar to the other two binary-query interactive segmentation algorithms \cite{ChenCC16,RupprechtPN15}. For comparison, at each interaction, all the three methods actively propose one seed to the user for acquiring a binary label.
We also compare with another three baselines. 
\djc{The first baseline `SwipeCut (1-layer)' uses the assessment function Eq.~(\ref{eq:funcScore}) on the superpixel-level graph. 
The second baseline `Random (1-layer)' just randomly proposes seeds on the superpixel-level graph. 
The third baseline `Random (2-layer)' randomly proposes seeds on the two-layer graph. In the baseline `SwipeCut (1-layer),' we also partition the superpixel-set $\mathcal{R}$ into a tree-set $\mathcal{T}$ for computing the proposal influence. However, its proposal confidence can only be calculated on the superpixel-level graph. }

\subsubsection{Segmentation Accuracy}
Note that, all variants of our method implement the segmentation on the superpixel-level graph. They differ only in the strategy of selecting query seeds. From Fig.~\ref{fig:comparison2} we can see that selecting the query seed in the two-layer graph is better than selecting the query seed in the single-layer graph. \djc{Notice that our approach `SwipeCut (1-layer),' which uses both color and texture features, just performs marginally better than \cite{ChenCC16} and \cite{RupprechtPN15}. However, our approach `SwipeCut (2-layer),' which uses both features and the different graph structure, has a noticeable improvement in segmentation accuracy. Therefore, we think that the improvement mainly comes from the use of the two-layer graph.} This is because the redundant query seeds (satisfying the label consistency assumption) are greatly suppressed.

The comparisons on our algorithm `SwipeCut (2-layer)' and the two previous methods of Chen~\textit{et al.}~\cite{ChenCC16} and Rupprecht~\textit{et al.}~\cite{RupprechtPN15} also show that our algorithm performs significantly better on those three datasets. The results imply that our approach is better than the existing methods on selecting the most informative query seed.

\subsubsection{Response Time}
The average response time per iteration of our method is less than $0.002$ seconds, which is far less than \cite{RupprechtPN15} ($< 1$ second) and slightly more than \cite{ChenCC16} ($<0.001$ seconds). The computation bottleneck of Rupprecht~\textit{et al.} is MCMC sampling, which is to approximate the image segmentation probability. The computation cost of Chen~\textit{et al.} is quite low, but it is outperformed by our approach on segmentation accuracy. Notice that the preprocessing step for over-segmentation and for building MST takes about $0.7$ seconds. However, it only needs to be done once before interaction and thus the efficiency of the entire algorithm would not be degraded. The time measurement is done on an Intel i7 $3.40$ GHz CPU with 8GB RAM.

\begin{table*}
\caption{\label{tab:comTimeTab} The average response time (seconds) per round of different algorithms. The measurement is done on an Intel i7-4770 $3.40$ GHz CPU with 8GB RAM. The timing results are obtained using the MSRA dataset. }
\vspace{-3mm}
\normalsize
\begin{center}
\begin{tabular}{|c||c|c|c|c|c|c|}
    \hline
    \textbf{Algorithm} & LazySnapping & RandomWalks & InteractiveGraphCuts & GeodesicStar & OneCut & SwipeCut  \\
    \hline \hline
    \textbf{Seconds}   & 0.33 & 0.72 & 0.34 & 0.61 & 0.44 & 0.002   \\
    \hline
\end{tabular}
\end{center}
\end{table*}

\begin{figure*}[t]
    \centering
    \subfigure[{\tiny ECSSD - various number of seeds}]
        { \includegraphics[width=0.31\textwidth]{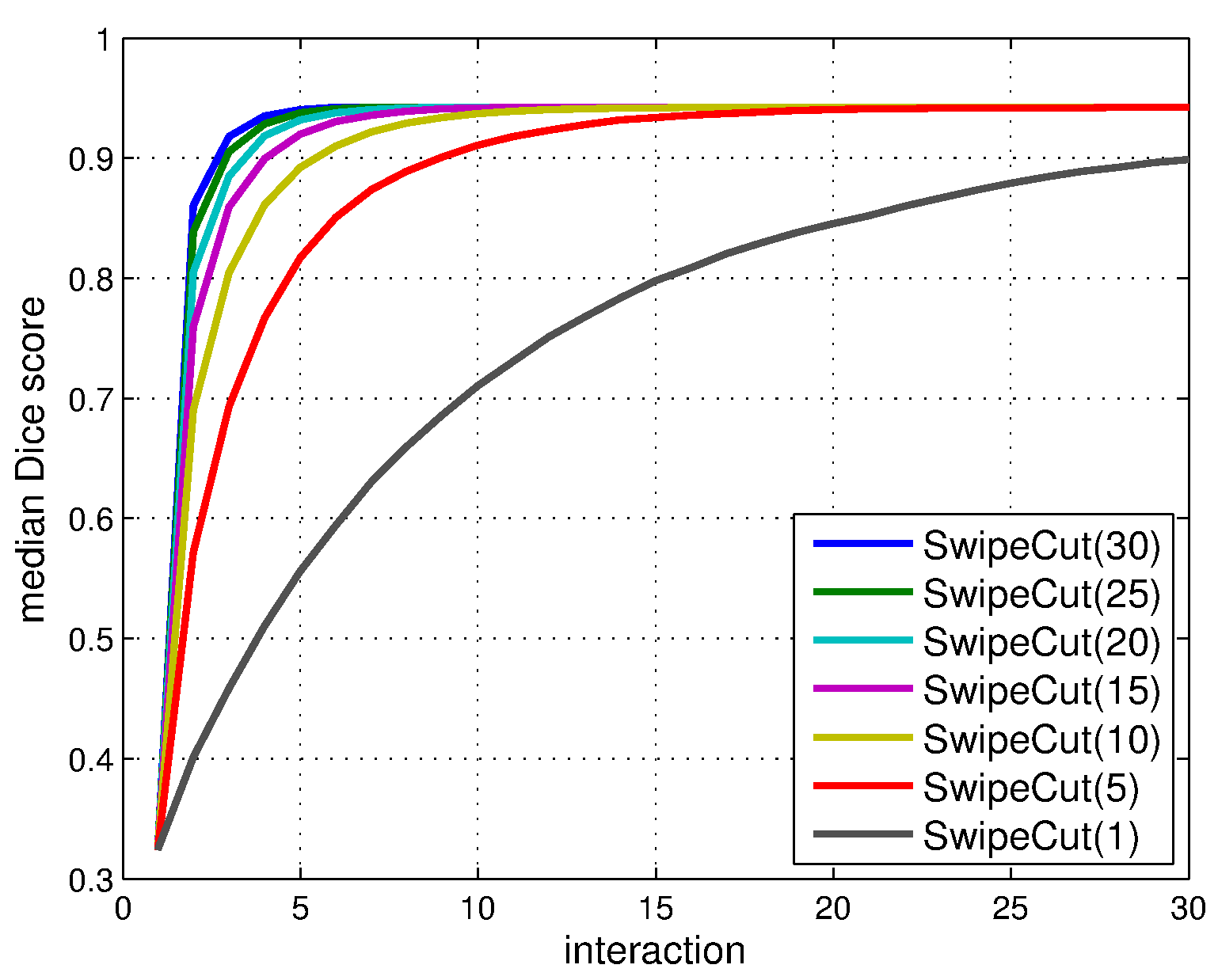} }
    \subfigure[{\tiny ECSSD - median Dice score}]
        { \includegraphics[width=0.31\textwidth]{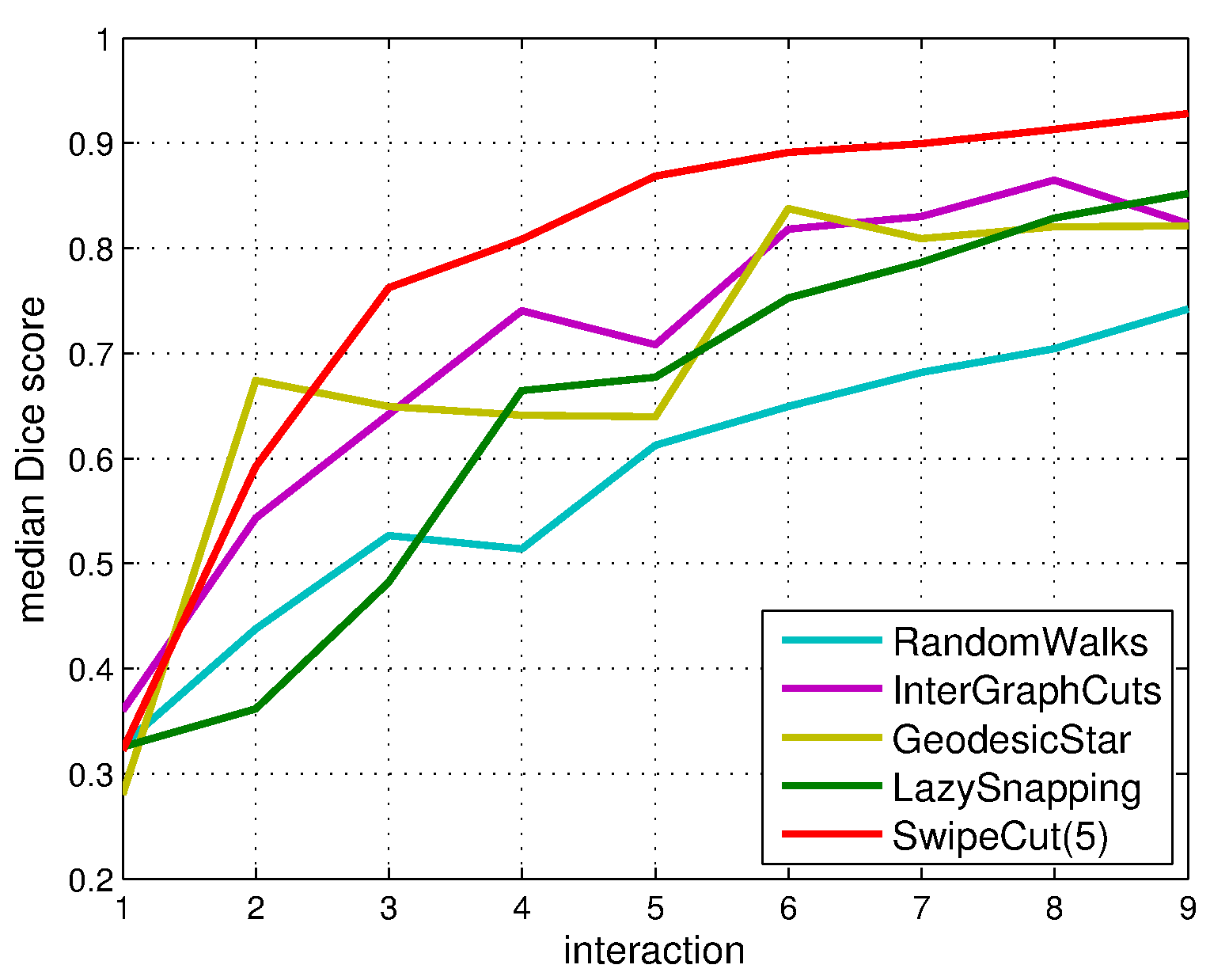} }
    \subfigure[{\tiny ECSSD - average time}]
        { \includegraphics[width=0.31\textwidth]{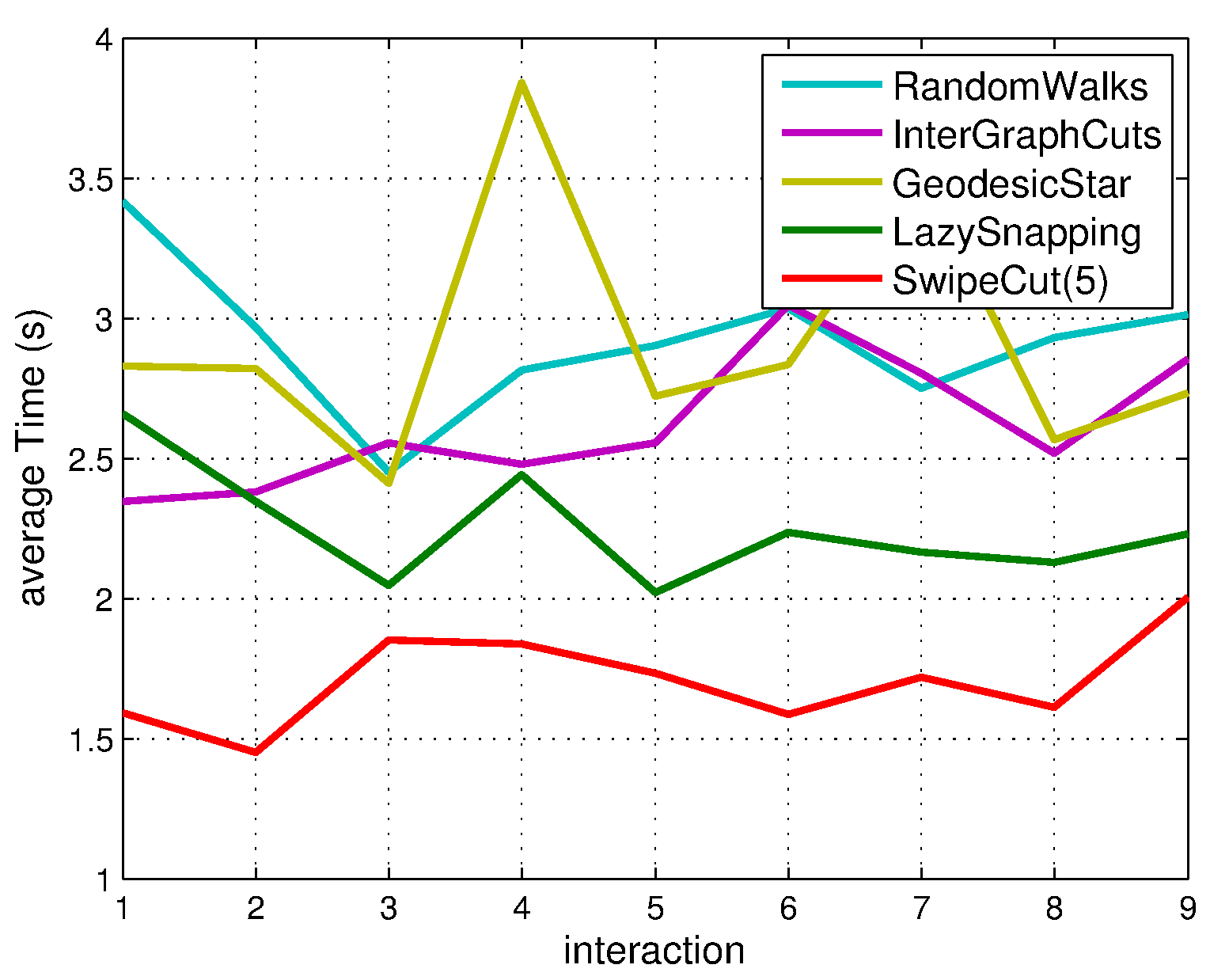} }
    \vspace{-3mm}
    \caption{\label{fig:comparison4} The evaluation of the number of query seeds per interaction on SwipeCut and the user study in comparison with different interactive segmentation algorithms using the ECSSD dataset. (a) The SwipeCut's performance on various number of query seeds per interaction. (b) The user study on the segmentation accuracy against the number of interactions. (c) The user study on the time cost against the number of interactions. The average time includes the user response time, I/O time, and computation time. }
\end{figure*}

\begin{figure*}[t]
    \centering    \vspace{-2mm}
        \begin{tabular}{m{1.4cm}m{0.95\textwidth}}
            & $\quad queries, GT \;\;\quad SwipeCut \qquad\quad\;\; LS \qquad\qquad\quad RW \qquad\qquad\; IGC \qquad\qquad\; GSC \qquad\qquad\; OC$  \\
            Round 5  & \multirow{5}{*}{ \includegraphics[width=0.87\textwidth]{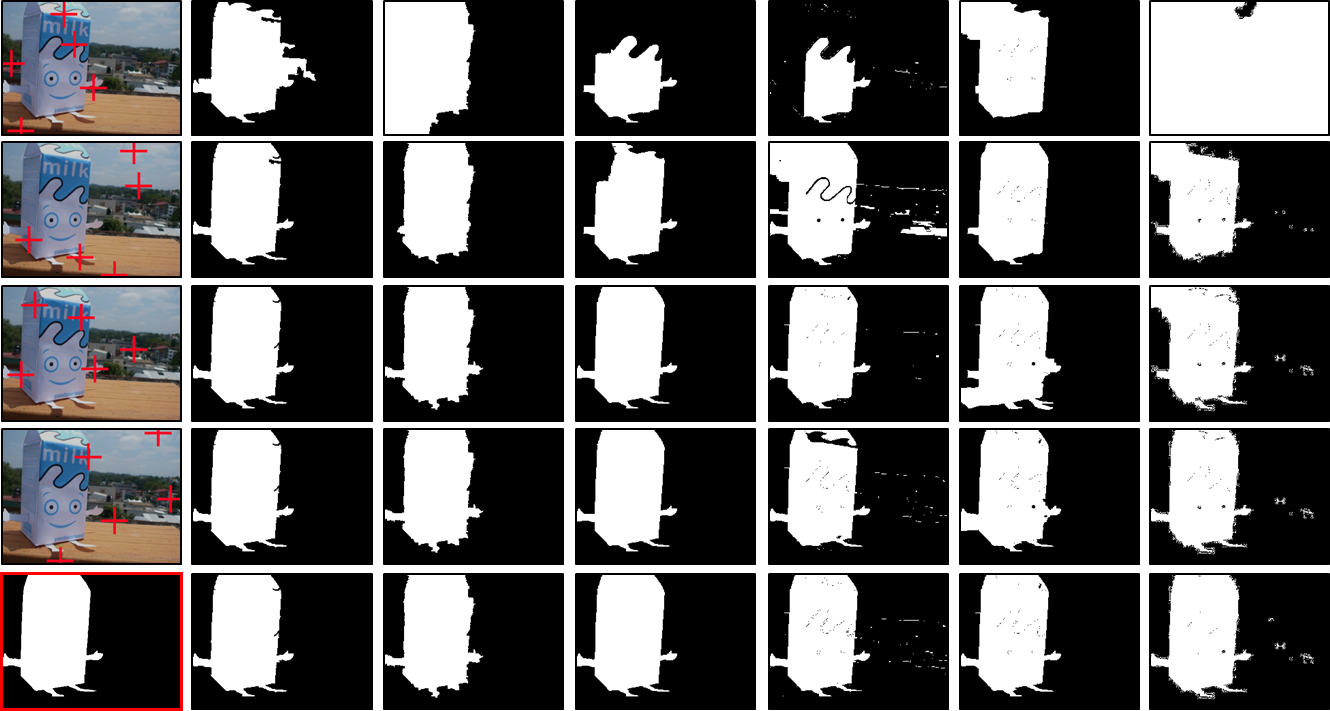} } \\[10mm]
            Round 10 &  \\[11mm]
            Round 15 &  \\[13mm]
            Round 20 &  \\[10mm]
            Round 30 &  \\[18mm]
            Round 5  & \multirow{5}{*}{ \includegraphics[width=0.87\textwidth]{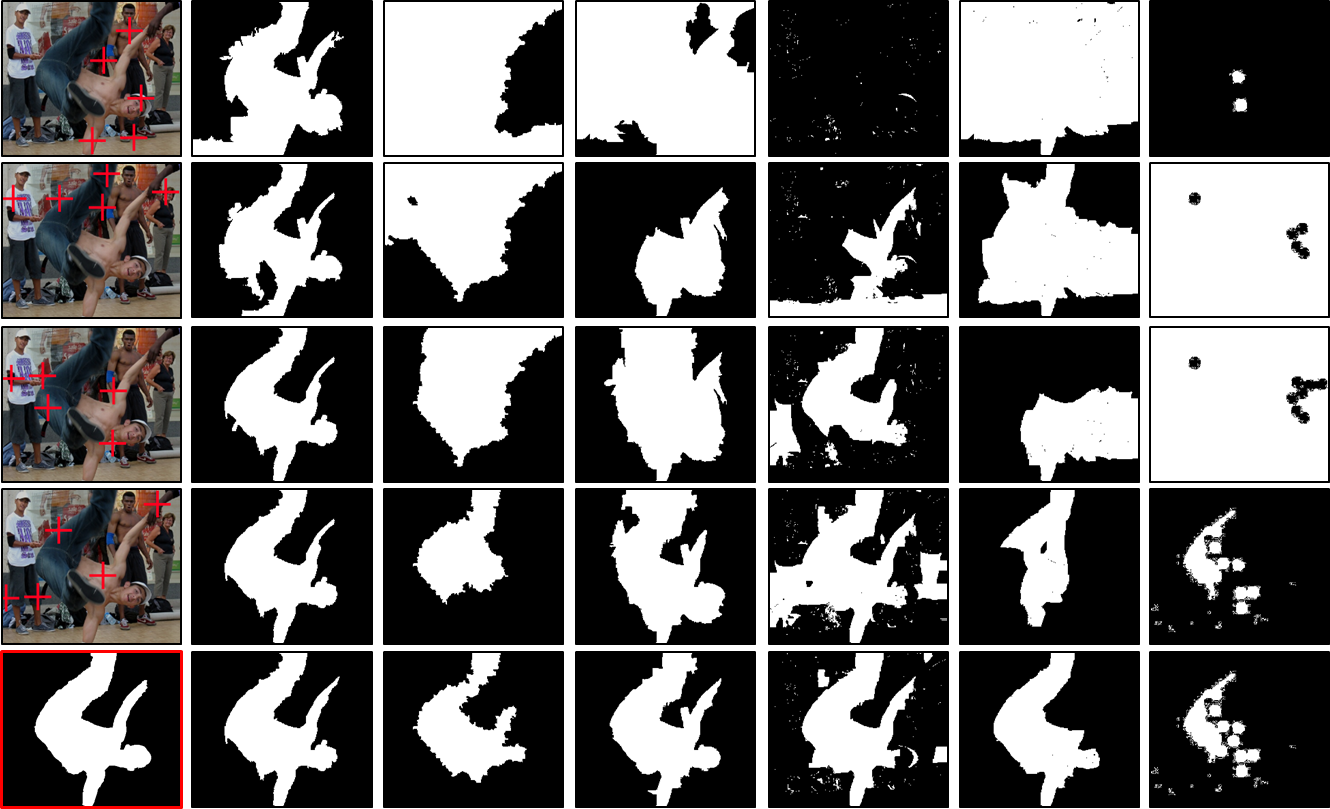} } \\[13mm]
            Round 10 &  \\[14mm]
            Round 15 &  \\[14mm]
            Round 20 &  \\[14mm]
            Round 30 &  \\[15mm]
        \end{tabular}
    \caption{\label{fig:visualCom}  Visualization of running examples of our approach. The top image set and the bottom image set respectively show the segmentation results using a simple background image and a cluttered background image. In the two image sets, the rows show the query seeds (in red crosses) and the segments from each method at the 5th, 10th, 15th, 20th, and 30th round. The image in red box shows the ground truth of the corresponding example image. Abbreviation of the methods in comparison: Lazy Snapping (LS), Random Walks (RW), Interactive Graph Cuts (IGC), Geodesic Star Convexity sequential (GSC), OneCut with seeds (OC).}
\end{figure*}

\subsection{Multiple Query Seeds Per Interaction}
We evaluate our approach on three datasets in this experiment. Our approach is compared with five state-of-the-art interactive segmentation algorithms, which are seed/scribble based algorithms listed as follows\footnote{The programs of Lazy Snapping are implemented by Gupta and Ramnath \url{http://www.cs.cmu.edu/~mohitg/segmentation.htm}.
The code of Random Walks is from \url{http://cns.bu.edu/~lgrady/software.html}
The programs of InterGraphCuts and GeodesicStar are from \url{http://www.robots.ox.ac.uk/~vgg/research/iseg/}. The code of OneCut is from \url{http://vision.csd.uwo.ca/code/}.}:
Lazy Snapping \cite{LiSTS04},
Random Walks \cite{Grady06},
Interactive Graph Cuts \cite{BoykovJ01},
Geodesic Star Convexity \cite{GulshanRCBZ10},
OneCut with seeds \cite{TangGVB13}.

Except our approach, the other five methods in this experiment do not have a seed-proposal mechanism. Therefore, we use the procedure in \cite{FengPCC16} to automatically synthesize the next seed position as a new user-input: In each round, each algorithm will \emph{ideally} select the centroid of the largest connected component among the exclusive-or regions between the current segmentation and the ground-truth segmentation, as is guided by an oracle. \djc{Notice that, our approach reasons out the seeds without the ground truth segmentation.}

Fig.~\ref{fig:comparison3} shows the experimental results on segmentation accuracy. The notation `$(\cdot)$' in Fig.~\ref{fig:comparison3} means the number of query seeds per interaction of the proposed approach, SwipeCut. The first line in Fig.~\ref{fig:comparison3} depicts the upper bound of our superpixel-level segmentation accuracy using 700 superpixels. Fig.~\ref{fig:comparison4}(a) shows the comparison of different settings on the number of query seeds per interaction.

It can be seen from Fig.~\ref{fig:comparison3} and Fig.~\ref{fig:comparison4}(a) that the version of multiple queries per iteration of our algorithm greatly boosts the segmentation accuracy. Collecting multiple labels per interaction makes our approach reach the segmentation-accuracy upper bound within fewer rounds.
In our approach, proposing five query seeds per interaction is sufficient to get better segmentation accuracy than other methods, while they all rely on the ideal oracle to select the seed for them in the experiment.
It is also worth emphasizing that the interaction mechanism of multiple query seeds per round is made viable owing to the specific formulation in Eq.~(\ref{eq:funcProblem}).
Furthermore, our seed-proposal and swipe-based mechanisms can be combined with other segmentation algorithms for acquiring multiple labels from the user.

We also present the average response time per round of different algorithms in Table.~\ref{tab:comTimeTab}. The additional visualization of running examples of the compared methods are shown in Fig.~\ref{fig:visualCom}. The advantage of our approach on interaction efficiency is clearly demonstrated.

\subsection{Evaluating User Interactions}
Fig.~\ref{fig:comparison4}(b) and Fig.~\ref{fig:comparison4}(c) depict the user study on segmentation efficiency of various interactive image segmentation algorithms. We ask ten users to segment twenty images from ECSSD dataset in ten rounds of interactions over five algorithms. Each image is shown with the corresponding ground truth to the users.
The goal of each user is to segment the given images and reproduce the corresponding ground truth as similar as possible. Our approach selects the query seeds for the user to swipe through, and all other algorithms show the user the segmentation derived from the user's previous annotations. Notice that, the users are not restricted to input merely the seed labels. Hence, each user can provide pixel-wise seed labels or longer line-drawing labels for guiding each segmentation algorithms. 

In the interactive image segmentation scenario, the results in Fig.~\ref{fig:comparison4}(b) and Fig.~\ref{fig:comparison4}(c) indicate that the users spend less time in average and achieve better segmentation accuracy via our algorithm. Therefore, the advantage of acquiring label using the `multiple-query-seeds with swipe gestures' strategy is evident, and our implementation approach carries out the strategy effectively and efficiently.


\section{Conclusion}
We have presented an effective approach to the interactive segmentation for small touchscreen devices. In our approach, the user only needs to swipe through the ROI-relevant query seeds, which is a common type of gesture for multi-touch user interface. Since the number of queries per interaction is constrained, the user has less burden to swipe trough the query seeds. Our label collection mechanism is flexible, and therefore other segmentation algorithms can also adopt our approach for acquiring multiple labels from the user in one round of interaction. \djc{Recently, deep learning based algorithms \cite{XuPCYH16,XuPCYH17,LiewWXOF17,ManinisCTG18} demonstrate good segmentation performance. The proposed interactive mechanism can be integrated with the deep features in addition to simple features like color and texture for improvements in segmentation accuracy.}
The experiments show that our interactive segmentation algorithm achieves the preferable properties of high segmentation accuracy and low response time, which are important for building user friendly applications of interactive segmentation.


%



\ifCLASSOPTIONcaptionsoff
  \newpage
\fi


\bibliographystyle{IEEEtran}
\bibliography{2017TIP}

\begin{thebibliography}{10}
\providecommand{\url}[1]{#1}
\csname url@samestyle\endcsname
\providecommand{\newblock}{\relax}
\providecommand{\bibinfo}[2]{#2}
\providecommand{\BIBentrySTDinterwordspacing}{\spaceskip=0pt\relax}
\providecommand{\BIBentryALTinterwordstretchfactor}{4}
\providecommand{\BIBentryALTinterwordspacing}{\spaceskip=\fontdimen2\font plus
\BIBentryALTinterwordstretchfactor\fontdimen3\font minus
  \fontdimen4\font\relax}
\providecommand{\BIBforeignlanguage}[2]{{%
\expandafter\ifx\csname l@#1\endcsname\relax
\typeout{** WARNING: IEEEtran.bst: No hyphenation pattern has been}%
\typeout{** loaded for the language `#1'. Using the pattern for}%
\typeout{** the default language instead.}%
\else
\language=\csname l@#1\endcsname
\fi
#2}}
\providecommand{\BIBdecl}{\relax}
\BIBdecl

\bibitem{BoykovJ01}
Y.~Boykov and M.~Jolly, ``Interactive graph cuts for optimal boundary and
  region segmentation of objects in {N-D} images,'' in \emph{ICCV}, 2001, pp.
  105--112.

\bibitem{DongSSY15}
X.~Dong, J.~Shen, L.~Shao, and M.~Yang, ``Interactive cosegmentation using
  global and local energy optimization,'' \emph{{IEEE} Trans. Image
  Processing}, vol.~24, no.~11, pp. 3966--3977, 2015.

\bibitem{FengPCC16}
J.~Feng, B.~Price, S.~Cohen, and S.~Chang, ``Interactive segmentation on rgbd
  images via cue selection,'' in \emph{CVPR}, 2016.

\bibitem{Grady06}
L.~Grady, ``Random walks for image segmentation,'' \emph{{IEEE} Trans. Pattern
  Anal. Mach. Intell.}, vol.~28, no.~11, pp. 1768--1783, 2006.

\bibitem{GulshanRCBZ10}
V.~Gulshan, C.~Rother, A.~Criminisi, A.~Blake, and A.~Zisserman, ``Geodesic
  star convexity for interactive image segmentation,'' in \emph{CVPR}, 2010,
  pp. 3129--3136.

\bibitem{WangHC14}
T.~Wang, B.~Han, and J.~P. Collomosse, ``Touchcut: Fast image and video
  segmentation using single-touch interaction,'' \emph{Computer Vision and
  Image Understanding}, vol. 120, pp. 14--30, 2014.

\bibitem{XuPCYH16}
N.~Xu, B.~L. Price, S.~Cohen, J.~Yang, and T.~S. Huang, ``Deep interactive
  object selection,'' in \emph{CVPR}, 2016, pp. 373--381.

\bibitem{LiewWXOF17}
J.~Liew, Y.~Wei, W.~Xiong, S.~Ong, and J.~Feng, ``Regional interactive image
  segmentation networks,'' in \emph{ICCV}, 2017, pp. 2746--2754.

\bibitem{ManinisCTG18}
K.~Maninis, S.~Caelles, J.~Pont{-}Tuset, and L.~V. Gool, ``Deep extreme cut:
  From extreme points to object segmentation,'' in \emph{CVPR}, 2018.

\bibitem{KassWT88}
M.~Kass, A.~P. Witkin, and D.~Terzopoulos, ``Snakes: Active contour models,''
  \emph{International Journal of Computer Vision}, vol.~1, no.~4, pp. 321--331,
  1988.

\bibitem{MortensenB95}
E.~N. Mortensen and W.~A. Barrett, ``Intelligent scissors for image
  composition,'' in \emph{SIGGRAPH}, 1995, pp. 191--198.

\bibitem{XianZCXD16}
M.~Xian, Y.~Zhang, H.~Cheng, F.~Xu, and J.~Ding, ``Neutro-connectedness cut,''
  \emph{{IEEE} Trans. Image Processing}, vol.~25, no.~10, pp. 4691--4703, 2016.

\bibitem{BadoualSUU17}
A.~Badoual, D.~Schmitter, V.~Uhlmann, and M.~Unser, ``Multiresolution
  subdivision snakes,'' \emph{{IEEE} Trans. Image Processing}, vol.~26, no.~3,
  pp. 1188--1201, 2017.

\bibitem{XuPCYH17}
N.~Xu, B.~L. Price, S.~Cohen, J.~Yang, and T.~S. Huang, ``Deep grabcut for
  object selection,'' in \emph{BMVC}, 2017.

\bibitem{ChengPZTR15}
M.~Cheng, V.~A. Prisacariu, S.~Zheng, P.~H.~S. Torr, and C.~Rother, ``Densecut:
  Densely connected crfs for realtime grabcut,'' \emph{Comput. Graph. Forum},
  vol.~34, no.~7, pp. 193--201, 2015.

\bibitem{RotherKB04}
C.~Rother, V.~Kolmogorov, and A.~Blake, ``"grabcut": interactive foreground
  extraction using iterated graph cuts,'' \emph{{ACM} Trans. Graph.}, vol.~23,
  no.~3, pp. 309--314, 2004.

\bibitem{ChenCC16}
D.~Chen, H.~Chen, and L.~Chang, ``Interactive segmentation from 1-bit
  feedback,'' in \emph{ACCV}, 2016.

\bibitem{RupprechtPN15}
C.~Rupprecht, L.~Peter, and N.~Navab, ``Image segmentation in twenty
  questions,'' in \emph{CVPR}, 2015, pp. 3314--3322.

\bibitem{BaiW14}
J.~Bai and X.~Wu, ``Error-tolerant scribbles based interactive image
  segmentation,'' in \emph{CVPR}, 2014, pp. 392--399.

\bibitem{SubrPSK13}
K.~Subr, S.~Paris, C.~Soler, and J.~Kautz, ``Accurate binary image selection
  from inaccurate user input,'' \emph{Comput. Graph. Forum}, vol.~32, no.~2,
  pp. 41--50, 2013.

\bibitem{BatraKPLC10}
D.~Batra, A.~Kowdle, D.~Parikh, J.~Luo, and T.~Chen, ``icoseg: Interactive
  co-segmentation with intelligent scribble guidance,'' in \emph{CVPR}, 2010,
  pp. 3169--3176.

\bibitem{FathiBRR11}
A.~Fathi, M.~Balcan, X.~Ren, and J.~M. Rehg, ``Combining self training and
  active learning for video segmentation,'' in \emph{BMVC}, 2011, pp. 1--11.

\bibitem{KowdleCGC11}
A.~Kowdle, Y.~Chang, A.~C. Gallagher, and T.~Chen, ``Active learning for
  piecewise planar 3d reconstruction,'' in \emph{CVPR}, 2011, pp. 929--936.

\bibitem{StraehleKKBDH12}
C.~N. Straehle, U.~K{\"{o}}the, G.~Knott, K.~L. Briggman, W.~Denk, and F.~A.
  Hamprecht, ``Seeded watershed cut uncertainty estimators for guided
  interactive segmentation,'' in \emph{CVPR}, 2012, pp. 765--772.

\bibitem{ArbelaezPBMM14}
P.~A. Arbel{\'{a}}ez, J.~Pont{-}Tuset, J.~T. Barron, F.~Marqu{\'{e}}s, and
  J.~Malik, ``Multiscale combinatorial grouping,'' in \emph{CVPR}, 2014, pp.
  328--335.

\bibitem{CarreiraS10}
J.~Carreira and C.~Sminchisescu, ``Constrained parametric min-cuts for
  automatic object segmentation,'' in \emph{CVPR}, 2010, pp. 3241--3248.

\bibitem{ManenGG13}
S.~Manen, M.~Guillaumin, and L.~J.~V. Gool, ``Prime object proposals with
  randomized prim's algorithm,'' in \emph{ICCV}, 2013, pp. 2536--2543.

\bibitem{UijlingsSGS13}
J.~R.~R. Uijlings, K.~E.~A. van~de Sande, T.~Gevers, and A.~W.~M. Smeulders,
  ``Selective search for object recognition,'' \emph{International Journal of
  Computer Vision}, vol. 104, no.~2, pp. 154--171, 2013.

\bibitem{WangZLZJW15}
C.~Wang, L.~Zhao, S.~Liang, L.~Zhang, J.~Jia, and Y.~Wei, ``Object proposal by
  multi-branch hierarchical segmentation,'' in \emph{CVPR}, 2015, pp.
  3873--3881.

\bibitem{XiaoLTLT15}
Y.~Xiao, C.~Lu, E.~Tsougenis, Y.~Lu, and C.~Tang, ``Complexity-adaptive
  distance metric for object proposals generation,'' in \emph{CVPR}, 2015, pp.
  778--786.

\bibitem{CarbonellG98}
J.~G. Carbonell and J.~Goldstein, ``The use of mmr, diversity-based reranking
  for reordering documents and producing summaries,'' in \emph{SIGIR}, 1998,
  pp. 335--336.

\bibitem{ChapelleWS02}
O.~Chapelle, J.~Weston, and B.~Sch{\"{o}}lkopf, ``Cluster kernels for
  semi-supervised learning,'' in \emph{NIPS}, 2002, pp. 585--592.

\bibitem{ZhouBLWS03}
D.~Zhou, O.~Bousquet, T.~N. Lal, J.~Weston, and B.~Sch{\"{o}}lkopf, ``Learning
  with local and global consistency,'' in \emph{NIPS}, 2003, pp. 321--328.

\bibitem{AchantaSSLFS12}
R.~Achanta, A.~Shaji, K.~Smith, A.~Lucchi, P.~Fua, and S.~S{\"{u}}sstrunk,
  ``{SLIC} superpixels compared to state-of-the-art superpixel methods,''
  \emph{{IEEE} Trans. Pattern Anal. Mach. Intell.}, vol.~34, no.~11, pp.
  2274--2282, 2012.

\bibitem{FelzenszwalbH04}
P.~F. Felzenszwalb and D.~P. Huttenlocher, ``Efficient graph-based image
  segmentation,'' \emph{International Journal of Computer Vision}, vol.~59,
  no.~2, pp. 167--181, 2004.

\bibitem{GrundmannKHE10a}
M.~Grundmann, V.~Kwatra, M.~Han, and I.~A. Essa, ``Efficient hierarchical
  graph-based video segmentation,'' in \emph{CVPR}, 2010, pp. 2141--2148.

\bibitem{KrahenbuhlK14}
P.~Kr{\"{a}}henb{\"{u}}hl and V.~Koltun, ``Geodesic object proposals,'' in
  \emph{ECCV}, 2014, pp. 725--739.

\bibitem{WangSP15}
W.~Wang, J.~Shen, and F.~Porikli, ``Saliency-aware geodesic video object
  segmentation,'' in \emph{CVPR}, 2015, pp. 3395--3402.

\bibitem{WeiWZS12}
Y.~Wei, F.~Wen, W.~Zhu, and J.~Sun, ``Geodesic saliency using background
  priors,'' in \emph{ECCV}, 2012, pp. 29--42.

\bibitem{GouldFK09}
S.~Gould, R.~Fulton, and D.~Koller, ``Decomposing a scene into geometric and
  semantically consistent regions,'' in \emph{ICCV}, 2009, pp. 1--8.

\bibitem{ShiYXJ16}
J.~Shi, Q.~Yan, L.~Xu, and J.~Jia, ``Hierarchical image saliency detection on
  extended {CSSD},'' \emph{{IEEE} Trans. Pattern Anal. Mach. Intell.}, vol.~38,
  no.~4, pp. 717--729, 2016.

\bibitem{AchantaHES09}
R.~Achanta, S.~S. Hemami, F.~J. Estrada, and S.~S{\"{u}}sstrunk,
  ``Frequency-tuned salient region detection,'' in \emph{CVPR}, 2009, pp.
  1597--1604.

\bibitem{LiuSZTS07}
T.~Liu, J.~Sun, N.~Zheng, X.~Tang, and H.~Shum, ``Learning to detect {A}
  salient object,'' in \emph{CVPR}, 2007.

\bibitem{VOC07}
M.~Everingham, L.~Van~Gool, C.~K.~I. Williams, J.~Winn, and A.~Zisserman, ``The
  {PASCAL} {V}isual {O}bject {C}lasses {C}hallenge 2007 {(VOC2007)}
  {R}esults,''
  http://www.pascal-network.org/challenges/VOC/voc2007/workshop/index.html.

\bibitem{FowlkesMM07}
C.~C. Fowlkes, D.~R. Martin, and J.~Malik, ``Local figure-ground cues are valid
  for natural images,'' \emph{Journal of Vision}, vol.~7, no.~8, pp. 1--9,
  2007.

\bibitem{Soensen48}
T.~S{\o}ensen, ``A method of establishing groups of equal amplitude in plant
  sociology based on similarity of species and its application to analyses of
  the vegetation on danish commons,'' \emph{Kongelige Danske Videnskabernes
  Selskab}, vol.~5, no.~4, pp. 1--34, 1948.

\bibitem{LiSTS04}
Y.~Li, J.~Sun, C.~Tang, and H.~Shum, ``Lazy snapping,'' \emph{{ACM} Trans.
  Graph.}, vol.~23, no.~3, pp. 303--308, 2004.

\bibitem{TangGVB13}
M.~Tang, L.~Gorelick, O.~Veksler, and Y.~Boykov, ``Grabcut in one cut,'' in
  \emph{{IEEE} International Conference on Computer Vision, {ICCV} 2013,
  Sydney, Australia, December 1-8, 2013}, 2013, pp. 1769--1776.

\end{thebibliography}
\end{document}